\documentclass[10pt]{article}
\usepackage[round]{natbib}

\usepackage{tabularx}
\usepackage[english]{babel}
\usepackage{color}
\usepackage{float}
\usepackage{amsmath}
\usepackage{multirow}
\usepackage{rotating}
\newcommand{\tabincell}[2]{\begin{tabular}{@{}#1@{}}#2\end{tabular}}

\usepackage[margin=1in]{geometry}

\usepackage[colorlinks,citecolor=red]{hyperref}
\hypersetup{colorlinks=true, linkcolor=blue, citecolor=blue}

\usepackage{graphicx}
\DeclareGraphicsExtensions{.pdf}
\usepackage[justification=Centering,singlelinecheck=false,caption=false,font=footnotesize]{subfig}

\usepackage{natbib}
\bibpunct{(}{)}{;}{a}{}{,}

\usepackage{graphicx}
\usepackage{url}
\usepackage{algorithm}
\usepackage{algorithmic}
\usepackage{subfig}
\usepackage{array}

\usepackage{multirow}
\usepackage{array}
\renewcommand{\arraystretch}{1.5}

\begin{document}

\title{\textbf{A Comprehensive Survey on Pose-Invariant\\Face Recognition}}

\author{Changxing Ding \hspace{1cm} Dacheng Tao \\
Centre for Quantum Computation and Intelligent Systems\\Faculty of Engineering and Information Technology\\University of Technology, Sydney\\81-115 Broadway, Ultimo, NSW\\Australia\\
Emails: chx.ding@gmail.com, dacheng.tao@uts.edu.au}

\date{15 March 2016}

\maketitle

\begin{abstract}
The capacity to recognize faces under varied poses is a fundamental human ability that presents a unique challenge for computer vision systems.
Compared to frontal face recognition, which has been intensively studied and has gradually matured in the past few decades,
pose-invariant face recognition (PIFR) remains a largely unsolved problem.
However, PIFR is crucial to realizing the full potential of face recognition for real-world applications,
since face recognition is intrinsically a passive biometric technology for recognizing uncooperative subjects.
In this paper, we discuss the inherent difficulties in PIFR and present a comprehensive review of established techniques.
Existing PIFR methods can be grouped into four categories, i.e., pose-robust feature extraction approaches,
multi-view subspace learning approaches, face synthesis approaches, and hybrid approaches.
The motivations, strategies, pros/cons, and performance of representative approaches are described and compared.
Moreover, promising directions for future research are discussed.

\end{abstract}

\section{Introduction} \label{introduction}
Face recognition has been one of the most intensively studied topics in computer vision for more than four decades.
Compared with other popular biometrics such as fingerprint, iris, and retina recognition,
face recognition has the potential to recognize uncooperative subjects in a non-intrusive manner.
Therefore, it can be applied to surveillance security, border control, forensics, digital entertainment, etc.
Indeed, numerous works in face recognition have been completed and great progress has been achieved,
from successfully identifying criminal suspects from surveillance cameras\footnote{http://ilinnews.com/armed-robber-identified-by-facial-recognition-technology-gets-22-years/}
to approaching human level performance on the popular Labeled Face in the Wild (LFW) database~\cite{taigman2014deepface,LFWTech}.
These successful cases, however, may be unrealistically optimistic as they are limited to near-frontal face recognition (NFFR).
Recent studies~\cite{li2014maximal,zhu2014multi} reveal that the best NFFR algorithms~\cite{chen2013blessing,taigman2009multiple,simonyan2013fisher,li2013probabilistic}
on LFW perform poorly in recognizing faces with large poses.
In fact, the key ability of pose-invariant face recognition (PIFR) desired by real-world applications remains largely unsolved,
as argued in a recent work~\cite{abiantunsparse}.

\begin{figure}
\centering
\includegraphics[width=0.9\linewidth]{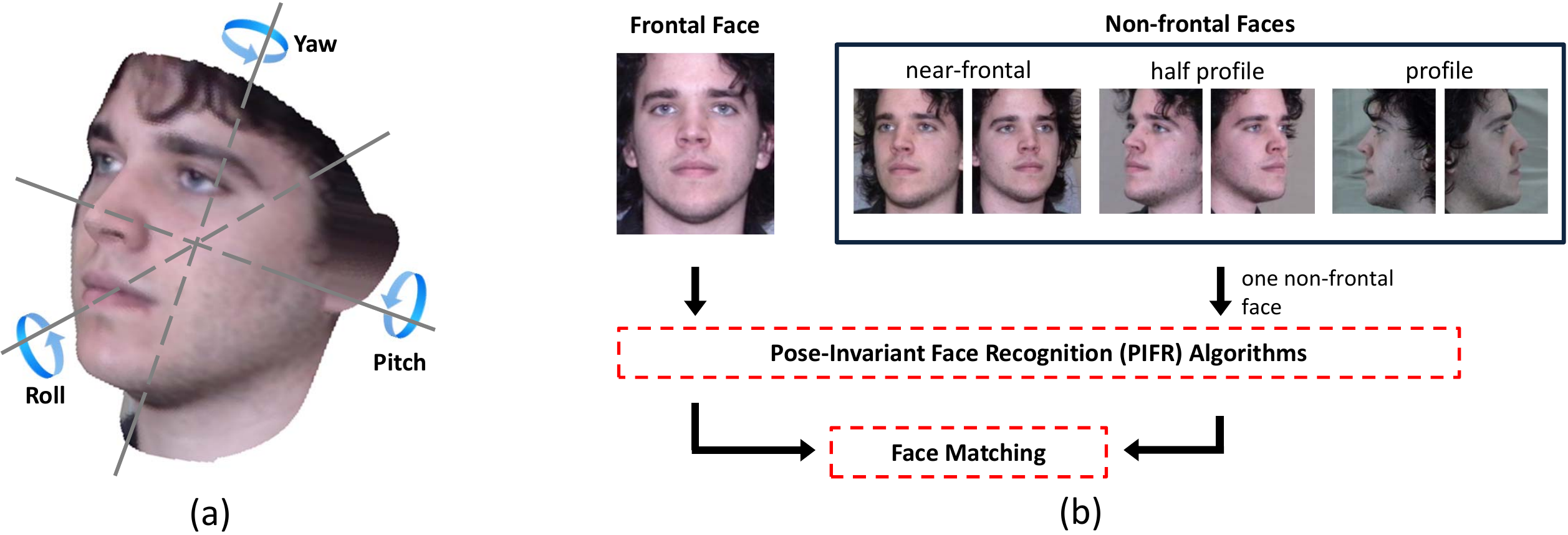}
\caption{(a) The three degrees of freedom of pose variation of the face, i.e., yaw, pitch, and roll.
(b) A typical framework of PIFR. Different from NFFR, PIFR aims to recognize faces captured under arbitrary poses.
}
\label{fig:PoseVariedFace}
\end{figure}

PIFR refers to the problem of identifying or authorizing individuals with face images captured under arbitrary poses,
as shown in Fig.~\ref{fig:PoseVariedFace}.
It is attracting more and more attentions, since face recognition is intrinsically a passive biometric technology for recognizing uncooperative subjects
and it is crucial to realize the full potential of face recognition technology for real-world applications.
For example, PIFR is important for biometric security control systems in airports, railway stations, banks,
and other public places where live surveillance cameras are employed to identify wanted individuals.
In these scenarios, the attention of the subjects is rarely focused on surveillance cameras
and there is a high probability that their face images will exhibit large pose variations.

\begin{figure}
\centering
\includegraphics[width=0.66\linewidth]{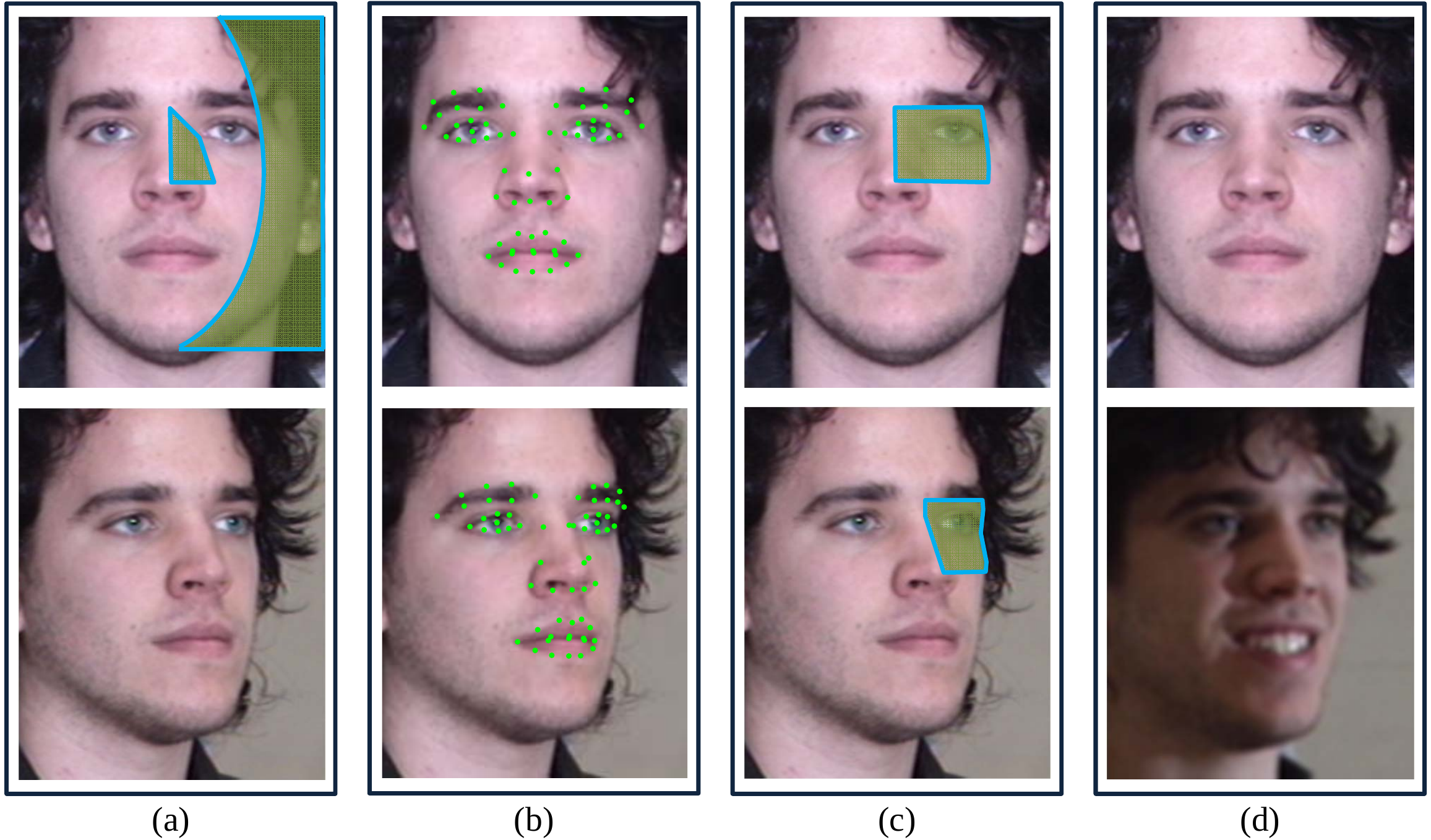}
\caption{The challenges for face recognition caused by pose variation.
(a) self-occlusion: the marked area in the frontal face is invisible in the non-frontal face;
(b) loss of semantic correspondence: the position of facial textures varies nonlinearly following the pose change;
(c) nonlinear warping of facial textures;
(d) accompanied variations in resolution, illumination, and expression.
}
\label{fig:PoseChallenge}
\end{figure}

The first explorations for PIFR date back to the early 1990s~\cite{brunelli1993face,pentland1994view,beymer1994face}.
Nevertheless, the substantial facial appearance change caused by pose variation continues to challenge the state-of-the-art face recognition systems.
Essentially, it results from the complex 3D structure of the human head. In detail, it presents the following challenges as illustrated in Fig.~\ref{fig:PoseChallenge}:

\begin{itemize}
\item  The rigid rotation of the head results in self-occlusion, which means there is loss of information for recognition.
\item  The position of facial texture varies nonlinearly following the pose change, which indicates the loss of semantic correspondence in 2D images.
\item  The shape of facial texture is warped nonlinearly along with the pose change, which causes serious confusion with the inter-personal texture difference.
\item  The pose variation is usually combined with other factors to simultaneously affect face appearance.
       For example, subjects being captured at a long distance tend to exhibit large pose variations, as they are unaware of the cameras.
       Therefore, low resolution as well as illumination variations occurs together with large pose variations.
\end{itemize}

For these reasons, the appearance change caused by pose variation often significantly surpasses the intrinsic differences between individuals.
In consequence, it is not possible or effective to directly compare two images under different poses, as in conventional face recognition algorithms.
Explicit strategies are required to bridge the cross-pose gap.
In recent years, a wide variety of approaches have been proposed which can be broadly grouped into the following four categories,
handling PIFR from distinct perspectives:

\begin{table}[h]
\renewcommand{\arraystretch}{1.3}
\caption{Taxonomy of Pose-Invariant Face Recognition Approaches}
\label{tab:Taxonomy}
\centering
\begin{tabular}{lp{7.5cm}}
\hline\hline
Category                                                        &Representative Works\\\hline
Pose-Robust Feature Extraction                                  &        \\
$\quad \ \bullet$ Engineered Features                           &Elastic Bunch Graph Matching~\cite{wiskott1997face}, Stereo Matching~\cite{castillo2009using} \\
$\quad \ \bullet$ Learning-based Features                       &Deep Neutral Networks~\cite{zhu2013deep,kan2014stacked}                                              \\
Multi-view Subspace Learning                                    &                                                                                                     \\
$\quad \ \bullet$ Linear Models                                 &CCA~\cite{li2009maximizing}, PLS~\cite{sharma2011bypassing}, GMLDA~\cite{sharma2012generalized},
                                                                 Tied Factor Analysis~\cite{prince2008tied} \\
$\quad \ \bullet$ Nonlinear Models                              &Kernel CCA~\cite{akaho2006kernel}, Deep CCA~\cite{andrew2013deep}        \\
Face Synthesis based on 2D Methods                              &    \\
$\quad \ \bullet$ 2D Pose Normalization                         &Piece-wise Warping~\cite{cootes2001active}, Patch-wise Affine Warping~\cite{ashraf2008learning}, MRFs~\cite{ho2013pose}        \\
$\quad \ \bullet$ Linear Regression Models                      &LLR~\cite{chai2007locally}, CCA~\cite{li2009maximizing}        \\
$\quad \ \bullet$ Nonlinear Regression Models                   &Deep Neutral Networks~\cite{zhu2013deep,kan2014stacked}        \\
Face Synthesis based on 3D Methods                              &    \\
$\quad \ \bullet$ Pose Normalization from Single Image          &3D Face Shape Model~\cite{jiang2005efficient,asthana2011pose}, Generic Elastic Models~\cite{heo20093d,heo2012gender}        \\
$\quad \ \bullet$ Pose Normalization from Multiple Images       &Frontal \& Profile Face Pairs~\cite{han20123d}, Stereo Matching~\cite{mostafa2012pose}        \\
$\quad \ \bullet$ 3D Modeling by Image Reconstruction           &3DMM~\cite{blanz1999morphable,aldrian2013inverse}        \\
Hybrid Methods                                                  &    \\
$\quad \ \bullet$ Feature \& Multi-view Subspace                &Block Gabor+PLS~\cite{fischer2012analysis}\\ 
$\quad \ \bullet$ Synthesis \& Multi-view Subspace              &PBPR-MtFTL~\cite{ding2015multi} \\
$\quad \ \bullet$ Synthesis \& Feature                          &Expert Fusion~\cite{kim2006design}, FR+FCN~\cite{zhu2014recover} \\\hline\hline
\end{tabular}
\end{table}

\begin{itemize}
\item  Those that extract pose-robust features as face representations, so that conventional classifiers can be employed for face matching.
\item  Those that project features of different poses into a shared latent subspace where the matching of the faces is meaningful.
\item  Those that synthesize face images from one pose to another pose,
       so that two faces originally in different poses can be matched in the same pose with traditional frontal face recognition algorithms.
\item  Those that combine two or three of the above techniques for more effective PIFR.
\end{itemize}

The four categories of approaches will be discussed in detail in later sections.
Inspired by~\cite{ouyang2014survey}, we unify the four categories of PIFR approaches in the following formulation.
\begin{eqnarray}
 &  & M\left[W^{a}F\left(S^{a}(\mathbf{I}_{i}^{a})\right),W^{b}F\left(S^{b}(\mathbf{I}_{j}^{b})\right)\right],
\label{eq:generalFramework}
\end{eqnarray}
where $\mathbf{I}_{i}^{a}$ and $\mathbf{I}_{j}^{b}$ stand for two face images in pose $a$ and pose $b$, respectively;
$S^{a}$ and $S^{b}$ are synthesis operations, after which the two face images are under the same pose;
$F$ denotes pose-robust feature extraction;
$W^{a}$ and $W^{b}$ correspond to feature transformations learnt by multi-view subspace learning algorithms;
and $M$ means a face matching algorithm, e.g., the nearest neighbor (NN) classifier.
It is easy to see that the first three categories of approach focus their effort on only one operation in Eq.~\ref{eq:generalFramework}.
For example, the multi-view subspace learning approaches provide strategies for determining the mappings $W^a$ and $W^b$;
the face synthesis-based methods are devoted to solving $S^{a}$ and $S^{b}$.
The hybrid approaches may contribute to two or more steps in Eq.~\ref{eq:generalFramework}.
Table~\ref{tab:Taxonomy} provides a list of representative approaches for each of the categories.

The remainder of the paper is organized as follows: Section 2 briefly reviews related surveys for face recognition.
Methods based on pose-robust feature extraction are described and analyzed in Section 3.
The multi-view subspace learning approaches are reviewed in Section 4.
Face synthesis approaches based on 2D and 3D techniques are respectively illustrated in Section 5 and Section 6.
The description of hybrid approaches follows in Section 7.
The performance of the reviewed approaches is evaluated in Section 8.
We close this paper in Section 9 by drawing some overall conclusions and making recommendations for future research.

\section{Related Works}
Numerous face recognition methods have been proposed due to the non-intrusive advantage of face recognition as a biometric technique.
Several surveys have been published. To name a few,
good surveys exist for illumination-invariant face recognition~\cite{zou2007illumination},
3D face recognition~\cite{bowyer2006survey}, single image-based face recognition~\cite{tan2006face},
video-based face recognition~\cite{barr2012face}, and heterogeneous face recognition~\cite{ouyang2014survey}.
There are also comprehensive surveys on various aspects of face recognition~\cite{zhao2003face}.
Of the existing works, the survey on face recognition across pose~\cite{zhang2009face}
that summarizes PIFR approaches before 2009 is the most relevant to this paper.
However, there are at least two reasons why a new survey on PIFR is imperative.

First, PIFR has become a particularly important and urgent topic in recent years
as the attention of the face recognition filed shifts from research into NFFR to PIFR.
The growing importance of PIFR has stimulated a more rapid developmental cycle for novel approaches and resources.
The increased number of PIFR publications over the last few years suggests new insights for PIFR, making a new survey for these methods necessary.

Second, the large-scale datasets for PIFR, i.e., Multi-PIE~\cite{gross2010multi} and IJB-A~\cite{klare2015pushing}, have only been established and made available in recent years,
creating the possibility of evaluating the performance of existing approaches in a relatively accurate manner.
In comparison, approaches reviewed in~\cite{zhang2009face} lack comprehensive evaluation as they use only small databases,
where the performance of many approaches have saturated.

This survey spans about 130 most innovative papers on PIFR, with more than 75\% of them published in the past seven years.
This paper categorizes these approaches from more systematic and comprehensive perspectives compared to~\cite{zhang2009face},
reports their performance on newly-developed large-scale datasets, explicitly analyzes the relative pros and cons of different categories of methods,
and recommends future development suggestions.
Besides, PIFR approaches that require more than two face images per subject for enrollment are not included in this paper,
as single image-based face recognition~\cite{tan2006face} dominates the research in the past decade.
Instead, we direct readers to~\cite{zhang2009face} the for good review on representative works~\cite{georghiades2001few,levine2006face,singh2007mosaicing}.

\section{Pose-robust feature extraction}
If the extracted features are pose-robust, then the difficulty of PIFR will be relieved.
Approaches in this category focus on designing a face representation that is intrinsically robust to pose variation
while remaining discriminative to the identity of subjects.
According to whether the features are extracted by manually designed descriptors,
or by trained machine learning models, the approaches reviewed in this section can be grouped into engineered features and learning-based features.

\subsection{Engineered Features}
\begin{figure}
\centering
\includegraphics[width=0.60\linewidth]{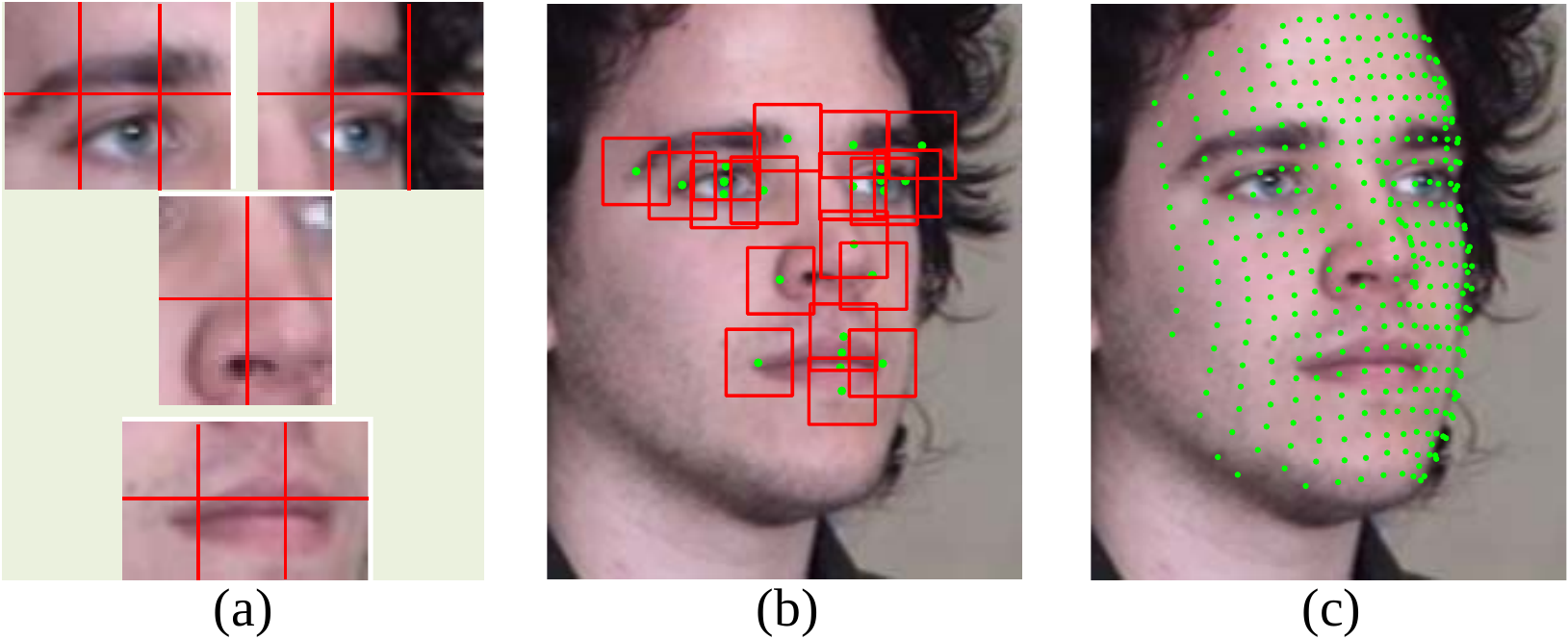}
\caption{Feature extraction from semantically corresponding patches or landmarks.
(a) Semantic correspondence realized in facial component-level~\cite{brunelli1993face,pentland1994view};
(b) Semantic correspondence by detecting dense facial landmarks~\cite{wiskott1997face,chen2013blessing,ding2014multi};
(c) Tight semantic correspondence realized with various techniques, e.g.,
3D face model~\cite{li2009maximizing,yi2013towards}, GMM~\cite{li2013probabilistic}, MRF~\cite{arashloo2011energy}, and stereo matching~\cite{castillo2011wide}.
}
\label{fig:EngineeredFeatures}
\end{figure}

Algorithms designed for frontal face recognition~\cite{turk1991eigenfaces,ahonen2006face} assume tight semantic correspondence between face images,
and they directly extract features from the rectangular region of a face image.
However, as shown in Fig.~\ref{fig:PoseChallenge}(b), one of the major challenges for PIFR is the loss of semantic correspondence in the face images.
To handle this problem, the engineered features reviewed in this subsection explicitly re-establish the semantic correspondence in the process of feature extraction,
as illustrated in Fig~\ref{fig:EngineeredFeatures}.
Depending on whether facial landmark detection is required,
approaches reviewed in this subsection are further divided into landmark detection-based methods and landmark detection-free methods.

\subsubsection{Landmark Detection-based Methods}
Early PIFR approaches~\cite{brunelli1993face,pentland1994view} realized semantic correspondence across pose at the facial component-level.
In~\cite{pentland1994view}, four sparse landmarks, i.e., both eye centers, the nose tip, and the mouth center, are first automatically detected.
Image regions containing the facial components, i.e., eyes, nose, and mouth, are estimated and the respective features are extracted.
The set of facial component-level features compose the pose-robust face representation.
Works that adopt similar ideas include~\cite{cao2010face, zhu2014recover}.

Better semantic correspondence across pose is achieved at the landmark-level.
\citet{wiskott1997face} proposed the Elastic Bunch Graph Matching (EBGM) model which iteratively deforms to detect dense landmarks.
Gabor magnitude coefficients at each landmark are extracted as the pose-robust feature.
Similarly,~\citet{biswas2013pose} described each landmark with SIFT features~\cite{lowe2004distinctive} and
concatenated the SIFT features of all landmarks as the face representation.
More recent engineered features benefit from the rapid progress in facial landmark detection~\cite{wang2014facial},
which makes dense landmark detection more reliable.
For example,~\citet{chen2013blessing} extracted multi-scale Local Binary Patterns (LBP) features from patches around 27 landmarks.
LBP features for all patches are concatenated to become a high-dimensional feature vector as the pose-robust feature.
A similar idea is adopted for feature extraction in~\cite{prince2008tied,zhangrandom}.

Intuitively, the larger the number of landmarks employed, the tighter semantic correspondence that can be achieved.
\citet{li2009maximizing} proposed the detection of a number of landmarks with the help of a generic 3D face model.
In comparison,~\citet{yi2013towards} proposed a more accurate approach by employing a deformable 3D face model with 352 pre-labeled landmarks.
Similar to~\cite{li2009maximizing}, the 2D face image is aligned to the deformable 3D face model using the weak perspective projection model,
after which the dense landmarks on the 3D model are projected to the 2D image.
Lastly, Gabor magnitude coefficients at all landmarks are extracted and concatenated as the pose-robust feature.

Concatenating the features of all landmarks across the face brings about highly non-linear intra-personal variation.
To relieve this problem,~\citet{ding2014multi} combined the component-level and landmark-level methods.
In their approach, the Dual-Cross Patterns (DCP)~\cite{ding2014multi} features of landmarks belonging to the same facial component are concatenated as the description of the component.
The pose-robust face representation incorporates a set of features of facial components.

While the above methods crop patches centered around facial landmarks,
\citet{fischer2012analysis} found that the location of the patches for non-frontal faces has a noticeable impact on the recognition results.
For example, the positions of patches around some landmarks, e.g., the nose tip and mouth corners,
for face images of extreme pose should be adjusted so that fewer background pixels are included.

\subsubsection{Landmark Detection-free Methods}
The accuracy and reliability of dense landmark detection are critical for building semantic correspondence.
However, accurate landmark detection in unconstrained images is still challenging.
To handle this problem,~\citep{zhao2009textural,liao2013partial,weng2013robust,li2015high}
proposed landmark detection-free approaches to extract features around the so-called facial keypoints.
For example,~\citet{liao2013partial} proposed the extraction of Multi-Keypoint Descriptors (MKD)
around keypoints detected by SIFT-like detectors.
The correspondence between keypoints among images is established via sparse representation-based classification (SRC).
However, the dictionary of SRC in this approach is very large, resulting in an efficiency problem in practical applications.
In comparison,~\citet{weng2013robust} proposed the Metric Learned Extended Robust Point set Matching (MLERPM) approach
to efficiently establish the correspondence between the keypoints of two faces.

Similarly,~\citet{arashloo2011pose} proposed an landmark detection-free approach based on Markov Random Field (MRF)
to match semantically corresponding patches between two images.
In their approach, the densely sampled image patches are represented as the nodes of the MRF model,
while the 2D displacement vectors are treated as labels.
The goal of the MRF-based optimization is to find the assignment of labels with minimum cost,
taking both translations and projective distortions into consideration.
The matching cost between patches can be measured from the gradient~\cite{arashloo2011energy}, or the gradient-based descriptors~\cite{arashloo2013efficient}.
The main shortcoming of this approach lies in the high computational burden in the optimization procedure,
which is accelerated by GPUs in their other works~\cite{arashloo2013efficient,rahimzadeh2014fast}.
For face recognition, local descriptors are employed to extract features from semantically corresponding patches~\cite{arashloo2013efficient,rahimzadeh2014class}.

Another landmark detection-free approach is the Probabilistic Elastic Matching (PEM) model proposed by~\citet{li2013probabilistic}.
Briefly, PEM first learns a Gaussian Mixture Model (GMM) from the spatial-appearance features~\cite{wright2009implicit}
of densely sampled image patches in the training set.
Each Gaussian component stands for patches of the same semantic meaning.
A testing image is represented as a bag of spatial-appearance features.
The patch whose feature induces the highest probability on each Gaussian component is found.
Concatenating the feature vectors of these patches forms the representation of the face.
Since all testing images follow the same procedure, the semantic correspondence is established.
They reported improved performance on LFW by establishing the semantic correspondence.
However, PEM has the disadvantage in efficiency, since the semantic correspondence is inferred through GMM.
As GMM only plays the role of a bridge to establish semantic correspondence between two images and the extracted features are still engineered,
we classify PEM as an engineered pose-robust feature.

To achieve pixel-wise correspondence between two face images under different poses,
~\citet{castillo2007using,castillo2009using,castillo2011wide} explored the stereo matching algorithm.
In their approach, four facial landmarks are first utilized to estimate the epipolar geometry of the two faces.
The correspondence between pixels of the two face images is then solved by a dynamic programming-based stereo matching algorithm.
Once the correspondence is known, normalized correlation based on raw image pixels is used to calculate the similarity score for each pair of corresponding pixels.
The summation of the similarity scores of all corresponding pixel pairs forms the similarity score of the image pair.
In another of their works~\cite{castillo2011trainable}, they replace raw image pixels with image descriptors to calculate the similarity of pixel pairs
and fuse the similarity scores using Support Vector Machine (SVM).

The engineered features handle the PIFR problem only from the perspective of establishing semantic correspondences,
which has clear limitations.
First, semantic correspondence may be completely lost due to self-occlusion in large pose images.
To cope with this problem,~\citet{arashloo2011energy} and~\citet{yi2013towards}
proposed the extraction of features only from the less-occluded half faces.
Second, the engineered features cannot relieve the challenge caused by the nonlinear warping of facial textures due to pose variation.
Therefore, the engineered features can generally handle only moderate pose variations.

\subsection{Learning-based Features}
The learning-based features are extracted by machine learning models that are usually pre-trained by multi-pose training data.
Compared with the engineered features, the learning-based features are better at handling the problem of self-occlusion and non-linear texture warping caused by pose variations.

\begin{figure}
\centering
\includegraphics[width=0.8\linewidth]{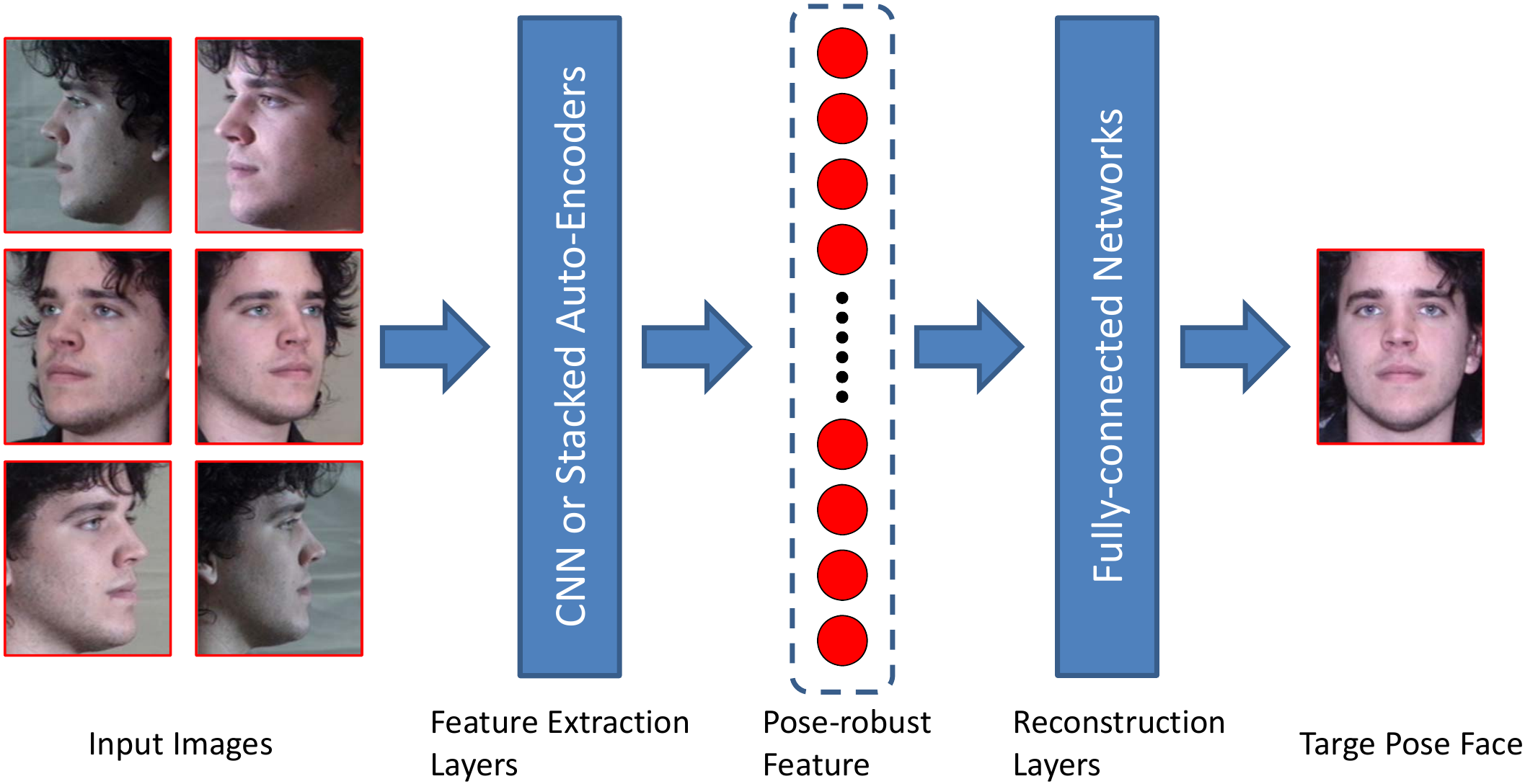}
\caption{The common framework of deep neural network-based pose-robust feature extraction methods~\cite{zhu2013deep,zhangrandom,kan2014stacked}.
}
\label{fig:LearningFeatures}
\end{figure}

Inspired by their impressive ability to learn high quality image representations,
neural networks have recently been employed to extract pose-robust features, as illustrated in Fig.~\ref{fig:LearningFeatures}.
\citet{zhu2013deep} designed a deep neural network (DNN) to learn the so called Face Identity-Preserving (FIP) features.
The deep network is the stack of two main modules: the feature extraction module and the frontal face reconstruction module.
The former module has three locally connected convolution layers and two pooling layers stacked alternately.
The latter module contains a fully-connected reconstruction layer.
The input of the model is a set of pose-varied images of an individual.
The output of the feature extraction module is employed to recover frontal faces through the latter module,
therefore the frontal face is saved as a supervised signal to train the network.
The logic of this method is that regardless of the pose of the input image,
the output of the reconstruction module is encouraged to be as close as possible to the frontal pose image of the subject.
Thus, the output of the feature extraction module must be pose-robust.
Due to the deep structure of the model, the network has millions of parameters to tune and therefore
requires a large amount of multi-pose training data.

Another contemporary work~\cite{zhangrandom} adopted a single-hidden-layer auto-encoder to extract pose-robust features.
Compared with~\cite{zhu2013deep}, the neural network built in~\cite{zhangrandom} is shallow because it contains only one single-hidden layer.
Like~\cite{zhu2013deep}, the input of the network is a set of pose-varied images of an individual,
but the target signal of the output layer is more flexible than~\cite{zhu2013deep},
i.e., it could be the frontal pose image of the identity or a random signal that is unique to the identity.
As argued by the authors, the target value for the output layer is essentially an identity representation which is not necessarily a frontal face,
therefore the vector that is represented by the neurons in the hidden layer can be used as the pose-robust feature.
Due to the shallow structure of the network, less amount of training data is required for training compared with~\cite{zhu2013deep}.
However, it may not extract as high-quality pose-robust feature as~\cite{zhu2013deep}.
To handle this problem, the authors proposed building multiple networks of exactly the same structure.
The input of the networks is the same, while their target values are different random signals.
In this way, parameters of the learnt network models are different,
and the multiple networks randomly encode a variety of information about the identity.
The final pose-robust feature is the vertical pile of the hidden layer outputs of all networks.

\citet{kan2014stacked} proposed the stacked progressive auto-encoders (SPAE) model to learn pose-robust features.
In contrast to~\cite{zhangrandom}, SPAE stacks multiple shallow auto-encoders, thus it is a deep network.
The authors argue that the direct transformation from the non-frontal face to the frontal face is a complex non-linear transform,
thus the objective may be trapped into local minima because of its large search region.
Inspired by the observations that pose variations change non-linearly but smoothly,
the authors proposed learning pose-robust features by progressive transformation from the non-frontal face to the frontal face through the stack of several shallow auto-encoders.
The function of each auto-encoder is to map the face images in large poses to a virtual view in slighter pose changes,
and meanwhile keep those images already in smaller poses unchanged.
In this way, the deep network is forced to approximate its eventual goal by several successive and tractable tasks.
Similar to~\cite{zhu2013deep}, the output of the top hidden layers of SPAE is used as the pose-robust feature.

The above three networks are single-task based, i.e., the extracted pose-robust features are required to reconstruct the face image under a single target pose.
In comparison,~\citet{jung2015rotating} designed a series interconnection network which includes a main DNN and an auxiliary DNN.
The pose-robust feature extracted by the main DNN is required not only to reconstruct the face image under the target pose,
but also recover the original input face image with the auxiliary DNN.
With the multi-task strategy, the identity-preserving ability of the extracted pose-robust features is observed to be enhanced compared with single-task based DNN.

Apart from the deep neural networks, a number of other machine learning models are utilized to extract pose-robust features.
For example, kernel-based models, e.g., Kernel PCA~\cite{liu2004gabor} and Kernel LDA~\cite{kim2006design,huang2007choosing,tao2006asymmetric,tao2007general,tao2009geometric},
were employed to learn nonlinear transformation to a high-dimensional feature space where faces of different subjects are assumed to be linearly separable,
despite of pose variation.
However, this assumption may not necessarily hold in real applications~\cite{zhang2009face}.
Besides, it has been shown that the coefficients for some face synthesis models~\cite{chai2007locally,li2012coupled,blanz2003face}
which will be reviewed in Sections 5 and 6 can be regarded as pose-robust features for recognition.
Their common shortcoming is that they suffer from statistical stability problems in image fitting due to the complex variations that appear in real images.

Another group of learning-based approaches is based on the face-similarity of one face image to $N$ template subjects~\cite{muller2007similarity,schroff2011pose,liao2013learning}.
Each of the template subjects has a number of pose-varied face images.
The pose-robust representation of the input image is also $N$ dimensional.
In~\cite{liao2013learning}, the $k$th element of the representation measures the similarity of the input image to the $k$th template subject.
This similarity score is obtained by first computing the convolution of the input image's low-level features with those of the $k$th template subject,
and then pooling the convolution results.
It is expected that the pooling operation will lead to robustness to the nonlinear pose variations.
In comparison,~\citet{schroff2011pose} proposed the Doppelg{\"a}nger list approach to sort the template subjects according to their similarity to the input face image.
The sorted Doppelg{\"a}nger list is utilized as the pose-robust feature of the input image.
Besides,~\citet{kafai2014reference} proposed the Reference Face Graph (RFG) approach to measure the discriminative power of each of the template subjects.
The similarity score of the input image to each template subject is modified by weighting the discriminative power of the template subject.
Their experiments demonstrate that performance is improved using this weighting strategy.
Compared with~\cite{zhu2013deep,zhangrandom,kan2014stacked},
the main advantage of the two approaches~\cite{liao2013learning,kafai2014reference} is that they have no free parameters.

\subsection{Discussion}
The engineered features achieve pose robustness by re-establishing the semantic correspondence between two images.
The semantic correspondence cannot handle the challenge of self-occlusion or nonlinear facial texture warping caused by pose variation.
The learning-based features compensate for this shortcoming by utilizing non-linear machine learning models, e.g., deep neural networks.
These machine learning models may produce higher quality pose-robust features,
but this is usually at the cost of massive labeled multi-pose training data,
which is not necessarily available in practical applications~\cite{liu2015Classification}.
The capacity of the learning-based features may be further enhanced by combining the benefit of semantic correspondence,
e.g., extracting features from semantically corresponding patches rather than the holistic face image.

\section{Multi-view Subspace Learning}
\begin{figure}
\centering
\includegraphics[width=0.60\linewidth]{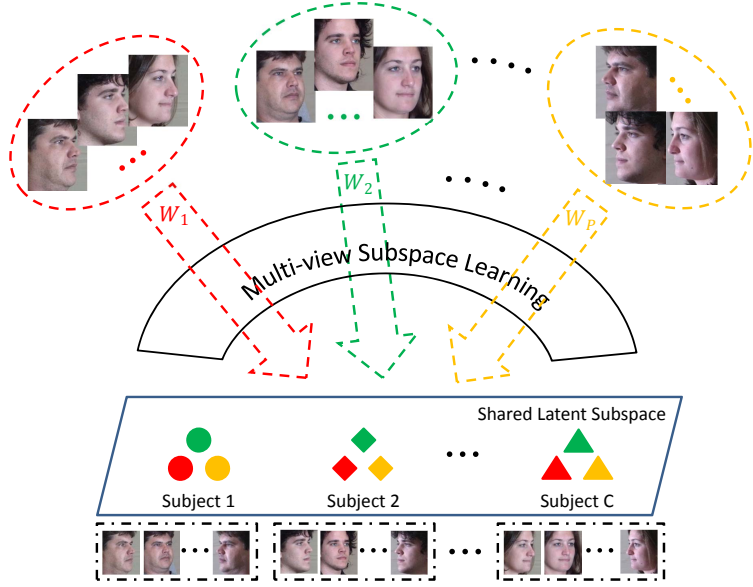}
\caption{The framework of multi-view subspace learning-based PIFR approaches~\cite{kan2012multi,prince2008tied,li2009maximizing,sharma2012generalized}.
The continuous pose range is divided into $P$ discrete pose spaces,
and pose-specific projections, i.e., $W_{1},W_{2},\dotsb,W_{P}$, to the latent subspace are learnt.
Approaches reviewed in this section differ in the optimization of the projections.
}
\label{fig:MultiViewSubSpace}
\end{figure}

The pose-varied face images are distributed on a highly nonlinear manifold~\cite{tenenbaum2000global},
which greatly degrades the performance of traditional face recognition models that are based on the single linear subspace assumption~\cite{turk1991eigenfaces}. %
The multi-view subspace learning-based approaches reviewed in this section tackle this problem by dividing the nonlinear manifold into a discrete set of pose spaces
and regard each pose as a single view,
and pose-specific projections to a latent subspace shared by different poses are subsequently learnt~\citep{kim2003discriminant,prince2005creating}.
Since the images of one subject are captured under different poses of the same face, they should be highly correlated in this subspace;
therefore face matching can be performed due to feature correspondence.
According to the properties of the models used, the approaches reviewed in this section are divided into linear models and nonlinear models.
An illustration of the multi-view subspace learning framework is shown in Fig~\ref{fig:MultiViewSubSpace}.

\subsection{Linear Models}
\subsubsection{Discriminative Linear Models}
\citet{li2009maximizing} proposed learning the multi-view subspace by exploiting Canonical Correlation Analysis (CCA).
The principle of CCA is to learn two projection matrices, one for each pose, to project the samples of the two poses into a common subspace,
where the correlation between the projected samples from the same subject is maximized.
Formally, given $N$ pairs of samples from two poses:
$\{(x_{11}, x_{21}), (x_{12}, x_{22}),\dotsb,(x_{1N}, x_{2N})\}$,
where $x_{pi}\in R^{d_{p}}$ represents the data of the $p^{th}$ pose from the $i^{th}$ pair, and $1\le p\le 2$, $1\le i\le N$.
It is required that the two samples in each pair belong to the same subject.
Two matrices $X_{1}=[x_{11}, x_{12},\dotsb,x_{1N}]$ and $X_{2}=[x_{21},x_{22},\dotsb,x_{2N}]$ are defined to represent the data from the two poses, respectively.
Two linear projections $w_{1}$ and $w_{2}$ are learnt for $X_{1}$ and $X_{2}$, respectively,
such that the correlation of the low-dimensional embeddings $w_{1}^{T}X_{1}$ and $w_{2}^{T}X_{2}$ is maximized:
\begin{equation}
\begin{array}{cl}
&\max\limits_{w_{1},w_{2}}corr[w_{1}^{T}X_{1}, w_{2}^{T}X_{2}]\\              
&s.t.\ \|w_{1}\|=1, \|w_{2}\|=1.\\    
\end{array}
\label{E:CCA}
\end{equation}
By employing the Lagrange multiplier, the above problem can be solved by the generalized eigenvalue decomposition method.
Since the projection of face vectors by CCA leads to feature correspondence in the shared subspace,
the subsequent face recognition can be conducted.

Considering the fact that CCA emphasizes only the correlation but ignores data variation in the shared subspace,
which may affect face recognition performance,
~\citet{sharma2011bypassing} and~\citet{li2011cross}
proposed the use of Partial Least Squares (PLS) to learn the multi-view subspace for both poses.
Formally, PLS finds the linear projections $w_{1}$ and $w_{2}$ such that
\begin{equation}
\begin{array}{cl}
&\max\limits_{w_{1},w_{2}}cov[w_{1}^{T}X_{1}, w_{2}^{T}X_{2}]\\              
&s.t.\ \|w_{1}\|=1, \|w_{2}\|=1.\\    
\end{array}
\label{E:PLS}
\end{equation}

Recall that the relation between correlation and covariance is as follows,
\begin{equation}
\begin{array}{cl}
corr[w_{1}^{T}X_{1}, w_{2}^{T}X_{2}]=\frac{cov[w_{1}^{T}X_{1}, w_{2}^{T}X_{2}]}{std(w_{1}^{T}X_{1})std(w_{2}^{T}X_{2})},
\end{array}
\label{E:CorrCov}
\end{equation}
where $std(\cdot)$ stands for standard deviation.
It is clear that PLS tries to correlate the samples of the same subject as well as capture the variations present in the original data,
which helps to enhance the ability to differentiate the training samples of different subjects in the shared subspace~\cite{sharma2011bypassing}.
Therefore, better performance by PLS than CCA was reported in~\cite{sharma2011bypassing}.

In contrast to CCA and PLS, which can only work in the scenario of two poses,
~\citet{rupnik2010multi} presented the Multi-view CCA (MCCA) approach to obtain one common subspace for all $P$ poses available in the training set.
In MCCA, a set of projection matrices, one for each of the $P$ poses, is learnt based on the objective of maximizing the sum of all pose pair-wise correlations:

\begin{equation}
\begin{array}{cl}
&\max\limits_{w_{1},\dotsb,w_{P}}\sum_{i<j} corr[w_{i}^{T}X_{i}, w_{j}^{T}X_{j}]\\              
&s.t.\ \|w_{i}\|=1, i=1,\dotsb,P\\    
\end{array}
\label{E:MCCA}
\end{equation}

The aforementioned methods only concern pair-wise closeness in the shared subspace and do not make use of the label information.
In contrast to the above methods,
~\citet{sharma2012robust} presented a two-stage framework for multi-view subspace learning.
In this method, the first stage learns pose-specific linear projections using MCCA,
after which all training samples are projected to the common subspace, which is assumed to be linear.
The second stage learns discriminative projections in the shared subspace using Linear Discriminant Analysis (LDA).
The experimental results indicate that the two-stage approach consistently results in improvements to performance,
compared to the unsupervised methods.

\citet{sharma2012generalized} further proposed a general framework named Generalized Multiview Analysis (GMA) for multi-view subspace learning.
The contribution of GMA is two-fold. First, existing methods based on generalized eigenvalue decomposition, e.g., PCA, CCA, PLS, and MCCA, can be unified in this framework.
Second, existing single-view models can be extended to their multi-view versions under this framework.
For example, the LDA model can be extended to its multi-view counterpart, Generalized Multi-view LDA (GMLDA).
Technically, GMLDA makes a tradeoff between the discriminability within each pose and the correlation between poses:
\begin{equation}
\begin{array}{cl}
&\max\limits_{w_{1},\dotsb,w_{P}}\sum\limits_{i=1}^P\mu_{i}w_{i}^{T}S_{bi}w_{i}+\sum_{i<j}\lambda_{ij}w_{i}^{T}Z_{i}Z_{j}^{T}w_{j}\\
&s.t.\ \sum\limits_{i=1}^P\gamma_{i}w_{i}^{T}S_{wi}w_{i}=1,\\
\end{array}
\label{E:GMLDA}
\end{equation}
where $\mu_{i}$, $\lambda_{ij}$, and $\gamma_{i}$ are model parameters.
$S_{bi}$ and $S_{wi}$ are the between-class scatter matrix and the within-class matrix, respectively, for the $i^{th}$ pose.
Therefore, the first term in the objective function enhances the discriminative power within each pose.
$Z_{i}$ are defined as the matrix with columns that are class means for the $i^{th}$ pose,
and corresponding columns of $Z_{i}$ and $Z_{j}$ should be of the same subject.
In this way, inter-pose faces of the same subject are correlated and clustered in the shared subspace,
and the gap caused by pose variation is thus reduced.
The experimental results in~\cite{sharma2012generalized} reveal that GMLDA outperforms their previously proposed two-stage method~\cite{sharma2012robust}.
Works that adopt a similar idea to GMA includes~\cite{huang2013cross,guo2014milda} which formulate multi-view subspace learning methods in graph embedding frameworks.

Methods based on CCA or PLS usually require each training subject to have the same number of faces for all poses,
a condition which may not be satisfied in real applications.
GMLDA relieves this demanding requirement but it still requires that each pose pair $(Z_{i},Z_{j})$ in Eq.~\ref{E:GMLDA} has exactly the same training subjects.
To handle this problem,~\citet{kan2012multi} proposed the Multi-view Discriminant Analysis (MvDA) approach which utilizes all face images from all poses.
In contrast to GMLDA, MvDA builds a single between-class scatter matrix $S_{b}$ and a single within-class scatter matrix $S_{w}$
from both the inter-pose and intra-pose embeddings in the shared subspace:
\begin{equation}
\begin{array}{cl}
&S_{w} = \sum\limits_{i=1}^C\sum\limits_{j=1}^P\sum\limits_{k=1}^{n_{ij}}(y_{ij}^{k}-\mu_{i})(y_{ij}^{k}-\mu_{i})^{T},\\
&S_{b} = \sum\limits_{i=1}^{C}n_{i}(\mu_{i}-\mu)(\mu_{i}-\mu)^{T},
\end{array}
\label{E:MvDA}
\end{equation}
where $y_{ij}^{k}$ stands for low-dimensional embedding of the $k^{th}$ sample from the $j^{th}$ pose of the $i^{th}$ subject.
$\mu_{i}$ is the mean of the low-dimensional embeddings of the $i^{th}$ subject,
and $\mu$ is the mean of the low-dimensional embeddings of all $C$ subjects.
The objective of MvDA is similar to that of LDA, i.e., to maximize the ratio between $S_{b}$ and $S_{w}$.
The pose-specific projections are obtained by formulating the objective as the optimization of a generalized Rayleigh quotient.
As variations from both inter-pose and intra-pose faces are considered together in the objective function,
the authors argued that a more discriminative subspace can be learnt.

\subsubsection{Generative Linear Models}
In addition to the above discriminative models, generative models are also explored for multi-view subspace learning.
One typical model is the Tied Factor Analysis (TFA) approach proposed by~\citet{prince2008tied}.
The core of TFA is the assumption that there exists an idealized identity subspace
where each multi-dimensional variable $h_{i}$ represents the identity of one subject, regardless of the pose.
Suppose $x_{ij}^{k}$ stands for the $k^{th}$ sample from the $j^{th}$ pose of the $i^{th}$ subject,
then $x_{ij}^{k}$ is generated by the pose-specific linear transformations of $h_{i}$ by $W_{j}$ in addition to the offset $\mu_{j}$
and Gaussian noise $\epsilon_{ij}^{k}\sim \mathcal{G} (0,\Sigma_{j})$:
\begin{equation}
\begin{array}{cl}
x_{ij}^{k} = W_{j}h_{i}+\mu_{j}+\epsilon_{ij}^{k}.\\
\end{array}
\label{E:TFA}
\end{equation}

The model parameters $W_{j}$, $\mu_{j}$, and $\Sigma_{j}$ ($1\le j\le P$) are estimated from multi-pose training data
by using the Expectation-Maximization (EM) algorithm.
In the recognition step, TFA calculates the probability that two images under different poses are generated from the same identity vector $h_{i}$
under the linear transformation scheme.

\citet{cai2013regularized} proposed the Regularized Latent Least Square Regression (RLLSR) method, which is similar to Prince's work~\cite{prince2008tied}.
RLLSR is based on the same assumption as TFA, however, it reformulates Eq.~\ref{E:TFA} in a least square regression framework,
where the observed images $x_{ij}^{k}$ is treated as regressor and the identity variable $h_{i}$ as response.
The identity variables $h_{i}$ and the model parameters $W_{j}$ and $\mu_{j}$ are estimated by the RLLSR model with an alternating optimization scheme.
To overcome overfitting and enhance the generalization ability, proper regularizations are imposed on the objective function of RLLSR.
In the recognition phase, the cosine metric is adopted to measure the similarity between the estimated identity variables.

The assumption that a single identity variable $h_{i}$ can be used to faithfully generate all different images of a subject under a certain pose seems to be over-simplified.
It is well-known that the appearance of a subject changes substantially as a result of the rich variations in illumination and expression and so forth,
even if the subject stays in the same pose.
In other words, TFA may not be able to effectively handle the complex within-class variations that appear in each pose space.
To overcome this limitation,~\citet{li2012probabilistic} proposed the Tied Probabilistic Linear Discriminant Analysis (Tied-PLDA) approach.
Tied-PLDA is built on TFA but incorporates both within-class and between-class variations in the model:
\begin{equation}
\begin{array}{cl}
x_{ij}^{k} = W_{j}h_{i}+G_{j}w_{ij}^{k}+\mu_{j}+\epsilon_{ij}^{k},\\
\end{array}
\label{E:TiedPLDA}
\end{equation}
where the respective definitions of $x_{ij}^{k}$, $h_{i}$, $\mu_{j}$, and $\epsilon_{ij}^{k}$ are the same as those in TFA.
$W_{j}$ and $G_{j}$ are pose-specific transformations which account for between-class subspace and within-class subspace, respectively.
All images belonging to the $i^{th}$ subject share the same identity variable $h_{i}$,
but each image has a different variable $w_{ij}^{k}$ which represents the position in the within-class subspace.
As in TFA, Tied-PLDA optimizes the model parameters using the EM algorithm.
In the recognition step, it calculates the probability that two images are generated from the same identity vector $h_{i}$,
regardless of whether they are in the same pose.
As the face generation process is formulated in a more reasonable way by Tied-PLDA, better performance than TFA was reported~\cite{li2012probabilistic}.

\subsection{Nonlinear Models}
Appearance changes in face images are highly nonlinear due to the substantial local warp and occlusion caused by pose variation.
The representational ability of the linear methods is limited, thus these methods may not be able to convert data from different poses into an ideal common space.
From this perspective, nonlinear techniques are preferable.

A natural nonlinear extension of the introduced linear models is realized via the kernel technique,
such that the nonlinear classification problem in the original space is converted to be the linear classification problem in the higher dimensional kernel space.
For example,~\citet{akaho2006kernel} proposed the extension of CCA to Kernel CCA (KCCA).
\citet{sharma2012generalized} extended the GMLDA approach to Kernel GMLDA (KGMLDA).
Better performance of KCCA than CCA on the face data was observed by~\citet{wang2014deeply}.

\begin{figure}
\centering
\includegraphics[width=0.45\linewidth]{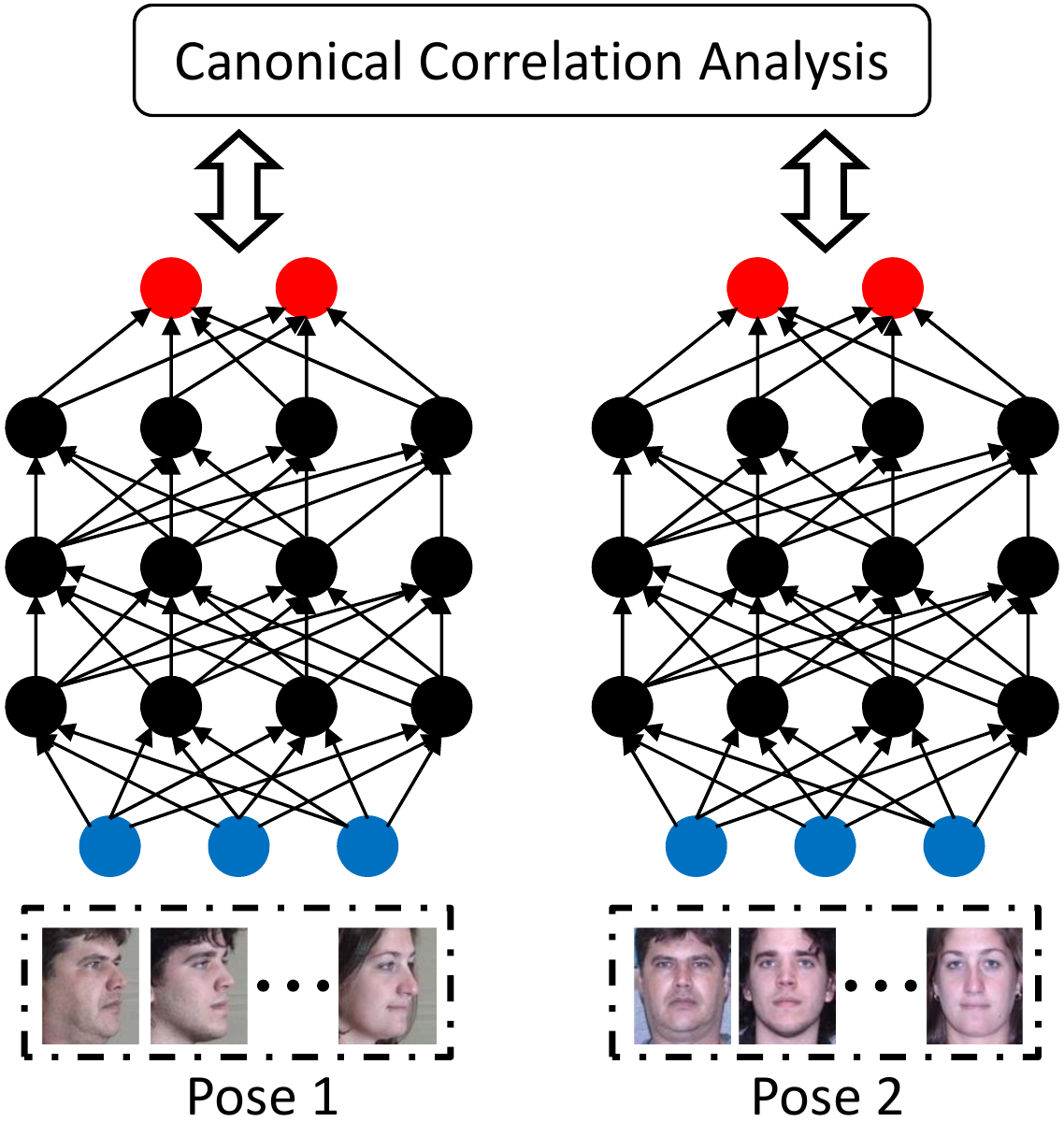}
\caption{The scheme of the Deep Canonical Correlation Analysis (DCCA) method~\cite{andrew2013deep}.
One network is learnt for each pose (view) and the outputs of the two networks are maximally correlated.
}
\label{fig:DCCA}
\end{figure}

Inspired by the ability of deep learning models to learn nonlinear representations,
a recent topic of interest has been to design multi-view subspace learning methods via deep structures.
\citet{andrew2013deep} proposed the Deep Canonical Correlation Analysis (DCCA) method,
which is another nonlinear extension of CCA.
In brief, DCCA builds one deep network for each of the two poses,
and the representations of the highest layer of the networks are constrained to be correlated,
as illustrated in Fig~\ref{fig:DCCA}.
Compared with KCCA, the training time of DCCA scales well with the size of the training set,
and if is not necessary to reference the training data in the testing stage.
It was reported in~\cite{andrew2013deep} that higher performance was achieved by DCCA than either CCA or KCCA.

DCCA is an unsupervised method, which means that it may not be suitable for classification tasks.
\citet{wang2014deeply} introduced the Deeply Coupled Auto-encoder Networks (DCAN) method to effectively employ the label information of the training data.
Similar to DCCA, DCAN also constructs one deep network for each of the two poses.
The two networks are discriminatively coupled with each other in every corresponding layer,
which enables samples from two poses to be projected to one common discriminative subspace.
The whole network is able to represent complex non-linear transformations as the number of layers increases;
therefore, the gap between the two poses narrows and the discriminative capacity of the common subspace is enhanced.
The authors reported better performance by DCAN than a number of linear methods, e.g., CCA, PLS, and MvDA,
and the nonlinear method KCCA, because of the nonlinear learning capacity of the deep networks.

\subsection{Discussion}
The multi-view subspace learning approaches reviewed in this section attempt to narrow the gap between different poses
by projecting their features to a common subspace with pose-specific transformations.
Among the existing techniques, linear models have the advantage in efficiency since the low-dimensional embeddings can be computed directly by matrix multiplication.
The capacity of the linear models is limited, however, as the appearance variations resulting from pose changes are intrinsically nonlinear.
The nonlinear techniques make up for this imperfection by learning nonlinear projections,
but at the cost of lower efficiency in model training or testing.
The deep learning-based nonlinear models also require larger training data.
The common shortcoming of the multi-view subspace learning methods
is that they depend on large training data which incorporate all the poses that might appear in the testing phase,
but the large amount of multi-pose training data might not be available in real-world applications.

\section{Face Synthesis based on 2D Methods}
Since directly matching two faces under different poses is difficult,
one intuitive method is to conduct face synthesis so that the two faces are transformed to the same pose,
allowing conventional face recognition algorithms to be used for matching.
Existing face synthesis methods for PIFR can be broadly classified into methods based on 2D techniques and methods based on 3D techniques,
depending on whether the synthesis is accomplished in the 2D domain or the 3D domain.
In this section, we review the 2D techniques which accomplish face synthesis directly in the 2D image domain.

\subsection{2D Pose Normalization}
\subsubsection{Piece-wise Warping}
\begin{figure}
\centering
\includegraphics[width=1.0\linewidth]{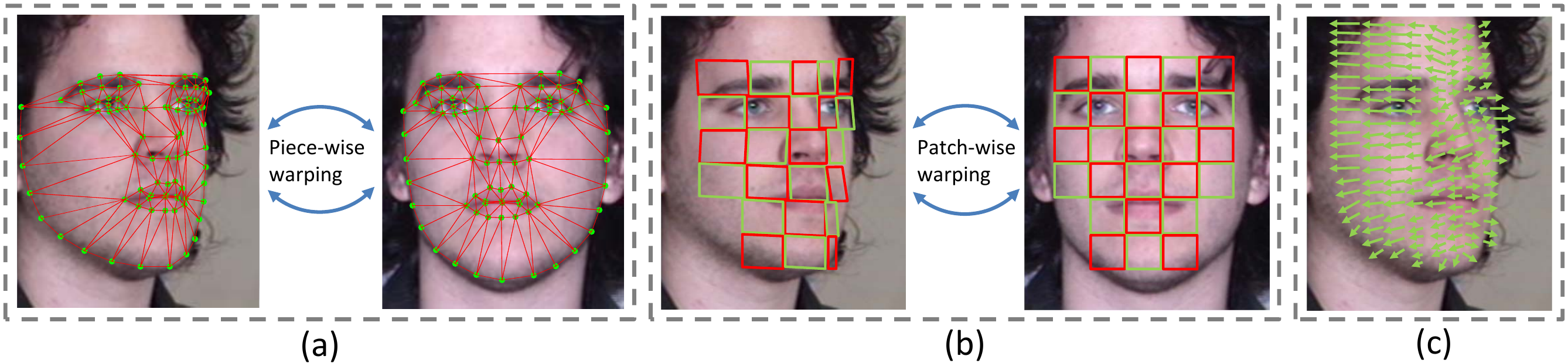}
\caption{Three main schemes of 2D-based pose normalization methods.
(a) Piece-wise warping; (b) Patch-wise warping; (c) Pixel-wise displacement.
}
\label{fig:TwoDSynthesis}
\end{figure}

The three main schemes of 2D pose normalization methods are illustrated in Fig.~\ref{fig:TwoDSynthesis}.
Early 2D pose normalization works for face synthesis are based on the piecewise warping strategy.
The piecewise warping approaches transform the shape of the face image in a piecewise manner to another specified pose.
Each piece refers to one triangle of the triangular mesh, which is generated by Delaunay triangulation of the dense facial landmarks on the face.
The warping transformation between each pair of triangle regions of the original image and the target image
can be an affine warping~\cite{gao2009pose} or a thin-plate splines-based warping~\cite{bookstein1989principal}.
The effects of the two warping strategies were compared by~\citet{gao2009pose}.

A number of works directly warp each face image to a uniform shape in the frontal pose with neutral expression~\cite{cootes2001active,gao2009pose}.
\citet{gonzalez2007toward} argued that warping all faces to the uniform shape is so strict
that the discriminative identity information for recognition may be lost.
They therefore collected a training set which incorporated the coordinates of dense facial landmarks.
A Point Distribution Model (PDM)~\cite{cootes1995active} was then built and the parameters that are assumed to control only pose variations were identified.
In the testing phase, the mesh of a testing image can be conveniently transformed to other poses by adjusting its pose parameters in PDM,
while the parameters controlling the identity and variations in expression stay the same.
In their experiments, the authors observed improved recognition performance over using a single uniform shape for face synthesis.
However, the identified pose parameters in PDM may not be well separated, i.e.,
they may also control the non-rigid transformation by expression or identity, resulting in error in the synthesized face images.

\citet{asthana2009learning} proposed the Gaussian Process Regression (GPR) model
to learn the correspondence between facial landmarks in the frontal and non-frontal poses offine.
In the testing phase, given a frontal face image and its facial landmarks, the landmark coordinates of non-frontal faces to be synthesized are predicted by GPR,
followed by warping the texture of the original frontal image.
In this way, the gallery set containing frontal faces is expanded with synthesized non-frontal faces and a probe image can be compared with gallery images in similar pose.
However, enlarging the gallery set reduces the efficiency of face recognition systems.
In another work~\cite{asthana2011pose}, the authors trained GPR to predict the landmark coordinates of the virtual frontal pose from the coordinates of a non-frontal face.
In this way, all the images are transformed to the frontal pose and compared without expanding the gallery set.

\citet{taigman2014deepface} proposed the inference of the landmark locations in the virtual frontal pose
for each non-frontal face with the help of a generic 3D face model.
First, the 3D face model is rotated to the pose of the 2D image by aligning the facial landmarks of both the model and the image.
The residuals between the landmarks of the 2D image and the projected locations of the 3D landmarks
are then added to the uniform shape in the frontal pose as compensation,
which is assumed to reduce the distortion to the identity caused by pose normalization.
In spite of the usage of 3D face model, pose normalization still happens in the 2D image domain.
Therefore, we categorize this approach as a 2D pose normalization technique.

The above approaches directly infer the personalized shape of a face in a novel pose.
\citet{berg2012tom} proposed another method of saving the identity information.
For each non-frontal image, triangulation is performed according to the generic landmark locations of all faces in this pose rather than the image's own facial landmarks.
The image is then transformed to the uniform shape of the frontal pose by piecewise affine warping.
In this way, a smaller distortion of the identity of the test image is achieved.
In the experiment on LFW, the authors reported better performance by the `identity-preserving' warping strategy than the uniform warping method.

Despite its simplicity, the piecewise warping method has obvious limitations.
First, the pose of the synthesized face cannot deviate too much from that of the original face because of the risk of causing severe degradation in image quality.
\citet{heo20123} investigated the ability of this method to handle yaw angle variation.
They concluded that if the yaw difference exceeds $\pm15^\circ$, the resulting warped images produce obvious stretching artifacts,
caused by over-sampling the texture in relatively small regions.
Second, the quality of the synthesized image depends heavily on the accuracy of the detection of each landmark as the warping is determined solely by the landmarks.
However, facial landmark detection continues to be a difficult problem for half-profile or profile images.

\subsubsection{Patch-wise Warping}
In contrast to the above works based on piecewise warping,~\citet{ashraf2008learning} modeled the face image as a collection of patches and
accomplished the reconstruction of the face image using a patch-wise strategy.
For each patch, they proposed the ``stack-flow'' approach for frontal patch synthesis from patches of a non-frontal pose.
In the training phase, the ``stack-flow'' method learns the optimal affine warp using the Lucas-Kanade (LK) algorithm
which aligns patches in the stack of the non-frontal faces to corresponding patches in the stack of frontal faces.
In the testing phase, each of the patches is warped to the frontal pose with the pre-learnt warp parameters, after which a frontal pose image is synthesized.
To stably learn the affine warp between the half-profile patches and the frontal patches,
the authors also proposed a composite strategy that successively learns the warp parameters between intermediate poses.
Due to the flexibility of the ``stack-flow'' approach, it is reported that it can cover a wider range of pose variations than piecewise warping approaches.

The patch-wise warps in the ``stack-flow'' approach are optimized individually for each patch,
without considering consistency at the overlapped pixels between two nearby patches.
To tackle this problem,~\citet{ho2013pose} proposed learning a globally optimal set of local warps for frontal face synthesis.
This method is composed of two main steps.
First, a set of candidate affine warps for each patch is obtained by aligning the patch to corresponding patches of a set of frontal training images,
using an improved LK algorithm~\cite{ashraf2010fast} that is robust to illumination variations.
Second, a globally optimal set of local warps $\{p_{i}\}_{i=1}^{N}$ is chosen from the candidate warps, one for each of the $N$ patches.
The problem is formulated into the following optimization problem:
\begin{equation}
\begin{array}{cl}
&\min\limits_{p_{1},\dotsb,p_{N}}\sum\limits_{i=1}^{N}E_{i}(p_{i})+\lambda\sum_{(i,j)\in \varepsilon}E_{ij}(p_{i},p_{j}),\\
\end{array}
\label{E:MRFSyn}
\end{equation}
where $E_{i}(p_{i})$ measures the cost of assigning the affine warp $p_{i}$ to the $i^{th}$ patch,
while $E_{ij}(p_{i},p_{j})$ is a smoothness term which measures the cost of inconsistency at the region of overlaps between the $i^{th}$ patch and $j^{th}$ patch.
$(i,j)\in \varepsilon$ if the two patches are four-connected neighbors.
Therefore, Eq.~\ref{E:MRFSyn} strikes a balance between the flexibility for each patch and the global consistency for all the patches.
Eq.~\ref{E:MRFSyn} is solved as a discrete labeling problem using MRF~\cite{ho2013pose}.

One common shortcoming for~\cite{ashraf2008learning} and~\cite{ho2013pose} is that they divide both frontal and non-frontal images into the same regular set of local patches.
This dividing strategy results in the loss of semantic correspondence for some patches when the pose difference is large,
as argued in~\cite{li2009maximizing},
therefore the learnt patch-wise affine warps may lose practical significance.
For various methods of detecting patches with semantic correspondence, please refer to Section 3 of this review.

\subsubsection{Pixel-wise Displacement}
The above piece-based or patch-based pose normalization methods cannot handle the local nonlinear warps that appear in each piece or patch.
\citet{beymer1995face} proposed the ``parallel deformation'' approach which
predicts the pixel-wise displacement between two poses.
This approach first establishes the dense pixel-wise semantic correspondence between images of different poses and subjects using the optical flow-based method.
The displacement fields containing dense pixel-wise displacement between two poses
from a set of training subjects are recorded as the template displacement fields.
Given a testing face image, its displacement field can be estimated by a linear combination of the template displacement fields.
With the estimated displacement field, the testing face image is deformed to the image under another pose.

The major drawback of the ``parallel deformation'' approach lies in the difficulty of establishing pixel-wise correspondence between images.
\citet{li2012morphable,li2014maximal} proposed the generation of the template displacement fields using images synthesized by a set of 3D face models.
The pixel-wise correspondence between the synthesized images can be easily inferred via the 3D model vertices,
therefore this approach implicitly utilizes 3D facial shape priors for pose normalization.
The authors also proposed the implicit Morphable Displacement Field (iMDF) method~\cite{li2012morphable}
and the Maximal Likelihood Correspondence Estimation (MLCE) method~\cite{li2014maximal}
to effectively estimate the convex combination of the template displacement fields.
In the testing phase, the 3D models are discarded and only the template displacement fields are utilized for face synthesis,
i.e., the face synthesis is accomplished in the 2D image domain.

\subsection{Linear Regression Models}
The same as the approaches in Section 4, methods described in this subsection divide the continuous pose space into a set of discrete pose segments.
Faces that fall into the same pose segment are assumed to have the same pose $p$.

The earliest work that formulated face synthesis as a linear regression problem was by~\citet{beymer1995face}.
Under the assumption of orthogonal projection and a condition of constant illumination,
the holistic face image is represented as the linear combination of a set of training faces in the same pose,
and the same combination coefficients are employed for face synthesis under another pose.
However, this approach requires dense pixel-wise correspondence between face images, which is challenging in practice.
Later works conducted face synthesis in a patch-wise strategy~\cite{chai2007locally}, reducing the difficulty in alignment.
Another advantage of the patch-based methods is that each patch can be regarded as a simple planar surface,
thus the transformation of the patches across poses can be approximated by linear regression models.

Inspired by Beymer's work~\cite{beymer1995face},~\citet{chai2007locally} proposed the Local Linear Regression (LLR) method for face synthesis.
LLR works on the patch level, based on the key assumption that the manifold structure of a local patch stays the same across poses.
Formally, suppose there exist two training matrices $D_{0}=[x_{01}, x_{02},\dotsb,x_{0n}]\in R^{d\times n}$ and $D_{p}=[x_{p1},x_{p2},\dotsb,x_{pn}]\in R^{d\times n}$,
whose columns are composed of the vectorized local patches in the frontal pose and non-frontal pose $p$, respectively.
Note that the corresponding columns (patches) in $D_{0}$ and $D_{p}$ are of the same subject.
In the testing time, given an image patch $x_{pt}$ in pose $p$, the first step of LLR is to predict $x_{pt}$ from the linear combination of columns in $D_{p}$,
where the combination coefficients $\alpha_{t}$ is computed by the least square algorithm:
\begin{equation}
\begin{array}{cl}
&\min\limits_{\alpha_{t}}\|x_{pt}-D_{p}\alpha_{t}\|^2.\\
\end{array}
\label{E:LLR1}
\end{equation}
The second step of LLR is to synthesize the frontal patch $x_{0t}$ with the learnt coefficients $\alpha_{t}$:
\begin{equation}
\begin{array}{cl}
x_{0t}=D_{0}\alpha_{t}.
\end{array}
\label{E:LLR2}
\end{equation}
By repeating the above process for all available patches in the testing image,
the final virtual frontal pose image can be obtained by overlapping all the predicted frontal patches.

In spite of its simplicity, LLR suffers from the overfitting problem in Eq.~\ref{E:LLR1},
which means the manifold structure learnt in pose $p$ may not faithfully represent the structure in another pose.
To relieve this problem, several improved approaches have been proposed by imposing regularization terms on Eq.~\ref{E:LLR1}.
These include the lasso regularization~\cite{li2012coupled,zhang2013pose}, the ridge regularization~\cite{li2012coupled},
local similarity regularization~\cite{hao2015unified},
and neighborhood consistency regularization~\cite{hao2015unified}.

Based on a similar assumption to the above works,
~\citet{yin2011associate} proposed the Associate-Predict model (APM) for frontal face synthesis.
The APM model has two steps for estimating the frontal pose patch from a non-frontal pose patch $x_{pt}$.
In the ``Associate'' step, $x_{pt}$ is associated with the most similar patch in $D_{p}$.
In the ``Predict'' step, the associated patch's corresponding patch in $D_{0}$ is directly utilized as a prediction for the frontal pose patch of $x_{pt}$,
and the predicted frontal patch is employed for the purpose of face matching.

Unlike many least square regression-based methods for face synthesis,
~\citet{li2009maximizing} proposed the formulation of CCA as a regressor for frontal face reconstruction.
In the training phase, CCA is employed to build the correlation-maximized subspace shared by the frontal and non-frontal patches,
as described in Section 4.
For frontal face synthesis, the non-frontal patches are first projected into the correlation-maximized subspace.
Then, ridge regression is employed to regress them into the frontal face space.
Overlapping all the synthesized frontal patches forms the required virtual frontal face.
As~\citet{li2009maximizing}'s approach is not based on such delicate assumption as that in~\cite{chai2007locally},
they report both higher synthesized image quality and higher recognition performance on the synthesized faces than~\cite{chai2007locally}.

In summary, although the linear regression-based methods are simple, they typically require a certain amount of multi-pose training data.
The regressed face images also suffer from the blurring effect and lose critical fine textures for recognition.
Moreover, for the approaches based on strict assumptions, e.g., identical local manifold structure across pose,
the synthesized faces are not guaranteed to be similar to the person appearing in the input image.

\subsection{Nonlinear Regression Models}
The appearance variation of the face image across poses is intrinsically nonlinear,
as a result of the substantial occlusion and nonlinear warp.
To synthesize face images of higher quality, nonlinear regression models have recently been introduced.

The four works reviewed in Section 3~\cite{zhu2013deep,zhangrandom,kan2014stacked,jung2015rotating} adopt neural networks as nonlinear regression models for 2D face synthesis,
inspired by their power to learn nonlinear transformations.
The common point for these four works is that they all first extract pose-robust features,
which are then utilized for frontal face recovery.
However, they have differences in structure.
\cite{zhu2013deep,jung2015rotating} adopted CNN to extract pose-robust features,
and face synthesis is accomplished via a full-connected reconstruction layer.
\citet{zhangrandom} employed single-hidden-layer auto-encoders for frontal face synthesis.
\citet{kan2014stacked} utilized stacked auto-encoders to recover the frontal face image from non-frontal input faces in a progressive manner,
thus reducing the difficulty of face synthesis for each auto-encoder.
Because of their deep structure,~\cite{zhu2013deep,kan2014stacked,jung2015rotating} may synthesize higher quality face images.

The common limitation of~\cite{zhu2013deep,zhangrandom,kan2014stacked} is that they are all deterministic networks that
recover face images of a fixed pose.
In comparison,~\cite{zhu2014multi,jung2015rotating} designed deep neural networks that can synthesize face images of varied poses.
For example,~\citet{zhu2014multi} proposed the Multi-View Perceptron (MVP) approach
to infer a wide range of pose-varied face images of the same identity, given a single input 2D face.
In brief, MVP first extracts pose-robust features with a similar structure as~\cite{zhu2013deep}.
The pose-robust feature is then combined with pose selective neurons to generate
the reconstruction feature, which is utilized to synthesize faces under the selected pose.
The face images of different poses are synthesized
with the varied outputs of the pose selective neurons.

\subsection{Discussion}
The 2D pose normalization-based methods conduct face synthesis by calculating
the warps across poses for each piece, patch, or pixel in the face image.
They require only limited or no training data, and they preserve the fine textures of the original image.
With the increase in pose difference,
however, the synthesized face image contains more significant stretching artifacts as a result of the dense sampling in relatively small regions.
The linear regression model-based methods compensate for this shortcoming but are usually based on strict assumptions about the local manifold structures across pose.
They handle a wider range of pose variations and require a moderate amount of training data.
However, these assumptions are not guaranteed to hold in practice.
The nonlinear regression model-based methods are the most powerful because they model the nonlinear appearance variation caused by pose change.
The major shortage is that they require substantial training data and considerable training time.
Moreover, the common shortcoming for both the linear or nonlinear regression methods is that the synthesized face image suffers from the blurring effect
and loses fine facial textures, e.g., moles, birthmarks, and wrinkles.
These peculiarities of textures are crucial for face recognition, as observed by~\citep{park2010face,li2015high}.

\section{Face Synthesis based on 3D Methods}
The human head is a complex non-planar 3D structure rotating in 3D space while the face image lies in the 2D domain.
The lack of one degree of freedom makes it difficult to conduct face synthesis using 2D techniques alone.
The methods reviewed in this section build a 3D model of the human head and then conduct face synthesis based on the 3D face models.
Face synthesis using 3D face models is one of the most successful strategies for PIFR.
The 3D methods can be classified into three subcategories, i.e.,
3D pose normalization from single image, 3D pose normalization from multiple images, and
3D modeling by image reconstruction.

\subsection{3D Pose Normalization from Single Image}
\begin{figure}
\centering
\includegraphics[width=0.6\linewidth]{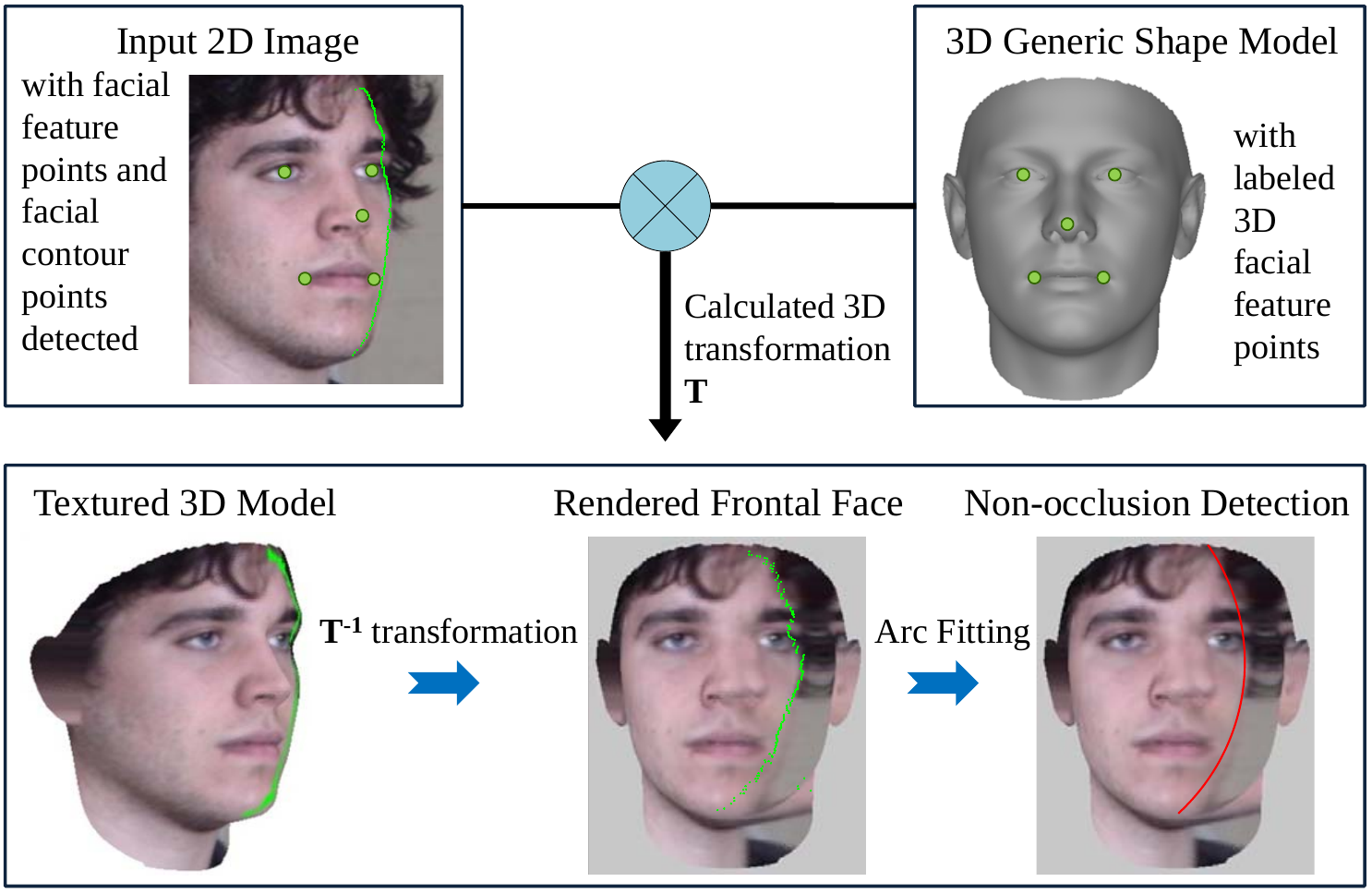}
\caption{The pipeline for 3D pose normalization from a single face image proposed in~\cite{ding2015multi}.
Face regions that are free from occlusion are detected and employed for face recognition.
}
\label{fig:3DPoseNorSingleIm}
\end{figure}

The 3D pose normalization approaches employ the 3D facial shape model as a tool to correct the nonlinear warp of facial textures appearing in the 2D image.
Like the 2D pose normalization methods reviewed in Section 5, they preserve the original pixel values of the input image.
As illustrated in Fig.~\ref{fig:3DPoseNorSingleIm}, the general principle is that the 2D face image is first aligned with a 3D face model,
typically with the help of facial landmarks~\cite{jiang2005efficient,hassner2013viewing}.
Then, the texture of the 2D image is mapped to the 3D model.
Lastly, the textured 3D model is rotated to a desired pose and a new 2D image in that pose is rendered.
Early approaches utilized simple 3D models, e.g., the cylinder model~\cite{gao2001fast}, the wire frame model~\cite{lee2003pose,zhang2006automatic},
and the ellipsoid model~\cite{liu2005pose}, to roughly model the 3D structure of the human head,
whereas newer approaches strive to build accurate 3D facial shape models.

\subsubsection{Normalization using PCA-based Face Models}
Most recent approaches build accurate 3D shape models by analyzing a set of aligned 3D facial scans~\cite{vetter1998synthesis,blanz2003face}.
Formally, suppose there are $N$ 3D meshes, each of which is composed of $n$ vertices,
and the full vertex-to-vertex correspondence between the $N$ meshes is established such that
vertices with the same index in each mesh have the same semantic meaning, e.g., the tip of the nose.
The geometry of the $i^{th}$ mesh can be represented as a shape vector that is composed of the 3D coordinates of the $n$ vertices:
\begin{equation}
\begin{array}{cl}
S_{i}=(x_{1}, y_{1}, z_{1}, x_{2},\dotsb,x_{n},y_{n},z_{n})^T\in R^{3n}.
\end{array}
\label{E:threeScan}
\end{equation}

The collection of $N$ meshes is represented as $S = \left[S_{1}, S_{2}, \dotsb, S_{N}\right]\in R^{3n\times N}$.
Since semantic correspondence has been established between the 3D scans,
it is reasonable to conduct Principle Component Analysis (PCA) on $S$ to model the variations of the 3D facial shape.
A new 3D face shape $S_{t}$ can then be represented as
\begin{equation}
\begin{array}{cl}
S_{t}=\bar{S}+A \alpha,
\end{array}
\label{E:pca3DShape}
\end{equation}
where $\bar{S}$ is the mean shape of $S$. $A\in R^{3n\times h}$ is the matrix stacked of $h$ eigenvectors
and $\alpha\in R^{h}$ contains the coefficients for the eigenvectors.

Given a 2D face image to be normalized, its landmarks are represented as
\begin{equation}
\begin{array}{cl}
L_{t}=(\hat{x_{1}}, \hat{y_{1}}, \hat{x_{2}},\dotsb, \hat{x_{m}}, \hat{y_{m}})^T\in R^{2m}.
\end{array}
\label{E:2DLandmark}
\end{equation}
Suppose the projection model of the camera is ${\bf P}$ which can be the orthogonal projection model or the perspective projection model.
${\bf \Omega}(S_{t})$ selects $m$ from all its $n$ vertices that correspond to the $m$ facial landmarks in the 2D image.
The rigid transformation ${\bf T}$ of the face and the shape parameter $\alpha$ are obtained by the following optimization,
\begin{equation}
\begin{array}{cl}
&\min\limits_{T,\alpha}\|L_{t}-{\bf{PT\Omega}}(S_{t})\|_{F}^{2}+\lambda \Phi(\alpha),\\
\end{array}
\label{E:TAlphaOpt}
\end{equation}
where $\Phi(\alpha)$ is a regularization term that penalizes the value of $\alpha$.
${\bf T}$ accounts for the translation, rotation, and scaling of the 3D face model $S_{t}$.
Lastly, the facial texture in the 2D image is mapped to the 3D face model with the shape parameter $\alpha$ and the rigid transformation ${\bf T}$.
The textured 3D model can be rotated to any desired pose, after which 2D face images under new poses can be rendered by computer graphics tools such as OpenGL.

Although theoretically simple, there are a number of factors to be considered in practical application.

First, \textbf{Generic 3D Model} vs \textbf{Personalized 3D Model}.
Typically, there is only one image for each gallery or probe subject.
Inferring a personalized 3D structure from a single 2D image is difficult,
therefore approaches that adopt the generic 3D model~\cite{ding2015multi,abiantunsparse,mostafa2012dynamic} simply calculate ${\bf T}$ by setting $\alpha$ as the zero vector,
i.e., ignoring the difference between subjects in the 3D structure.
This is reasonable to some extent since the difference between subjects in the 3D structure is minor.
There are also approaches that emphasize the importance of individual differences in facial structure
for face recognition~\cite{niinuma2013automatic,jo2015single}.
In this case, the shape parameter $\alpha$ and transformation parameter ${\bf T}$ are solved in an alternative optimization fashion.
In case of overfitting,~\citet{patel20093d} and~\citet{yi2013towards} proposed the imposition of regularization on the magnitude of $\alpha$.
Although appealing, these approaches rely on the accurate detection of dense facial landmarks and may suffer from statistical stability problems.
This indicates that different values of $\alpha$ are obtained from different images of the same subject,
which adversely affects face recognition, as argued by~\citet{hassner2015effective}.
Another practical limitation is observed by~\citet{jiang2005efficient},
who revealed that the popular 3D face databases contain 3D scans from only a small number of subjects,
e.g., 100 in the USF Human ID 3D database~\cite{blanz1999morphable}, 100 in BU-3DFE database~\cite{yin2008high}, and 200 in the Basel Face Model~\cite{paysan20093d}.
Therefore, the face space $A$ in Eq.~\ref{E:pca3DShape} spanned by these scans is quite limited and may not cover the rich variations appearing in testing images.

Second, \textbf{Accurate Correspondence}.
The accuracy of Eq.~\ref{E:TAlphaOpt} depends on the correct correspondence between the 2D landmarks and 3D vertices, i.e., they should be of the same semantic meaning.
However, the correspondence between the 2D landmarks on the facial contour and 3D model vertices are pose-dependent, as observed by~\citep{asthana2011fully}.
This is because the position of facial contour in the 2D image changes along with pose variations.
To handle this problem,~\citep{lee2012single,qu2014fast} proposed to detect and discard the moved landmarks.
~\citet{asthana2011fully} proposed the construction of a lookup table
which contains the manually labeled 3D vertices that correspond to the 2D facial landmarks under a set of discrete poses.
In the testing phase, the pose of the test image is first estimated and the nearest pose in the lookup table is determined.
The corresponding 3D model vertices recorded in the table are employed for Eq.~\ref{E:TAlphaOpt}.
In contrast to~\cite{asthana2011fully},~\citep{ding2012continuous,zhu2015high} proposed automatic approaches
to establish the pose-dependent correspondences online.
For example, in~\cite{ding2012continuous}, the pose of the test image is first estimated,
after which a virtual image under the same pose is rendered using a textured generic 3D model.
The facial landmarks of the virtual image are detected.
Then, the depth buffer at the position of facial landmarks is searched and corresponding 3D vertices are determined.
Compared with~\cite{asthana2011fully},~\citet{ding2012continuous}'s approach spares the need for tedious offline labeling work;
however, it relies on virtual image rendering and is thus less efficient in the testing phase.

Third, \textbf{Occlusion Detection}.
The 3D pose normalization approaches rely on texture mapping to synthesize face images in novel poses;
however, facial textures are incomplete for a non-frontal face due to self-occlusion.
When the 3D model is textured from the non-frontal face,
the vertices on the far-side of the 3D model cannot be correctly textured.
Therefore, when rendering new face images under the frontal or opposite poses,
the facial textures rendered by the occluded 3D vertices are not useful for recognition.
To distinguish the occluded and occlusion-free textures in the rendered face image,
~\citet{li2014maximal} proposed the use of pose-specific masks in the rendered image.
\citet{ding2012continuous} introduced the Z-buffer algorithm~\cite{van2014computer} to determine the visibility of each vertex on the 3D model.
Similarly,~\citet{abiantunsparse} proposed the inference of the visibility of each vertex
by comparing its normal direction and the viewing direction of the camera.
If the angle between the two directions is beyond a certain value, then the vertex is regarded as invisible.
Nevertheless, the above methods may not be accurate for occlusion detection because they assume accurate estimation for the pose and 3D facial structure.
Recently, inspired by the fact that the major boundary between the occluded and un-occluded regions is the facial contour,
~\citet{ding2015multi} proposed a robust algorithm to detect the facial contour of non-frontal face images.
As shown in Fig.~\ref{fig:3DPoseNorSingleIm}, the facial contour is projected to the rendered frontal face image and fitted with an arc,
which serves as the natural boundary between the occluded and un-occluded facial textures.
This strategy detects occlusion precisely and is free from accurate pose estimation.

Fourth, \textbf{Synthesis of the Occluded Texture}.
After occlusion detection as described above,
some approaches employ only the un-occluded textures for recognition~\cite{li2014maximal,ding2015multi},
while others synthesize the occluded textures such that a complete face image is obtained.
For example,~\citet{ding2012continuous} fill the occluded textures by copying the horizontally symmetrical textures.
\citep{hassner2015effective,zhu2015high} observed that this strategy may produce artifacts when the lighting conditions on both sides of the face are different,
and proposed compensation methods, respectively.
Also, the human face is asymmetrical, particularly when facial expressions or occlusion are involved,
as argued by~\citet{abiantunsparse}.
Instead,~\citet{abiantunsparse} proposed to fit the un-occluded pixels to a PCA model that is trained by the frontal faces.
\begin{equation}
\begin{array}{cl}
&\min\limits_{c}\|x'-{\bf{\Omega}}(Vc+m)\|_{F}^{2}+\lambda \|c\|_{1},\\
\end{array}
\label{E:OccFitting}
\end{equation}
where $x'$ is the vector containing the un-occluded pixels in the image.
$V$ and $m$ are computed by the PCA model with the frontal image training set.
$\bf{\Omega}$ is a selection matrix that selects the pixels corresponding to $x'$ from $Vc+m$.
The learnt PCA coefficients vector $c$ is employed for the reconstruction of the occluded pixels.

Fifth, \textbf{Synthesize Frontal Faces} vs \textbf{Non-frontal Faces}.
The most common setting for the PIFR research and application is that the gallery set is composed of controlled frontal faces,
while the probe set contains pose-varied faces.
Therefore, there are two distinct choices for face synthesis:
(i) Transforming each of the probe images to the frontal pose online~\cite{ding2015multi,best2014unconstrained}.
In this case, the traditional frontal face recognition algorithms are employed to match each probe image with the gallery images;
(ii) Transforming each of the gallery images to a discrete set of pose-varied images offline~\cite{niinuma2013automatic,hsu2013face}.
Given a probe image, its pose is first estimated and the gallery images of the nearest pose are selected for face matching.
This strategy is called the view-based face recognition~\cite{beymer1994face}.
Choice (i) has the advantage of handling continuous pose variations that appear in the probe set,
while Choice (ii) divides the pose space into discrete sections and is affected by the accuracy in pose estimation.
The latter choice is also less efficient as it stores a large number of gallery images in memory.
The advantage of the latter choice, however, lies in the fact that automatic 3D modeling from frontal images is typically easier than modeling from profile faces.
This is because dense facial landmark detection, which is important for 3D modeling, is usually unstable for profile faces due to severe self-occlusion.

Sixth, \textbf{Handling Expression Variations}.
Eq.~\ref{E:pca3DShape} does not take expression variations into consideration,
i.e., the face images are assumed to be of neutral expression. This impacts on recognition in two ways.
First, the rendered faces save the expression that appears in the original face image,
which may be different from the expression in the image to be compared.
To render faces with varied expressions,
~\citet{jiang2005efficient} proposed using the MEPG-4-based animation framework to
drive the textured 3D face model to present different expressions.
Second, the value of the PCA coefficient vector $\alpha$ in Eq.~\ref{E:pca3DShape} is contaminated by expression variations in the input image.
To handle this problem,~\citet{chu20143d} proposed expending Eq.~\ref{E:pca3DShape} as follows.
\begin{equation}
\begin{array}{cl}
S_{t}=\bar{S}+\left[A_{id} \ A_{exp}\right] \begin{bmatrix}\alpha_{id}\\ \alpha_{exp}\end{bmatrix},
\end{array}
\label{E:pca3DShapeExp}
\end{equation}
where $A_{id}$ and $A_{exp}$ are PCA models that account for identity and expression variations, respectively.
$A_{exp}$ can be trained on databases which contain a set of expressive 3D scans~\cite{yin2008high,cao2014facewarehouse}.
Correspondingly, $\alpha_{id}$ and $\alpha_{exp}$ stand for the identity and expression coefficient vectors, respectively.
After fitting an image and mapping the facial textures in the input image to the 3D model, the obtained $\alpha_{id}$ remains unchanged,
while $\alpha_{exp}$ is forced to be the coefficients of the neutral expression.
Therefore, the 3D model and the subsequently rendered images would be of neutral expression.
Although it is more powerful, the major drawback of Eq.~\ref{E:pca3DShapeExp} is that the separation between $\alpha_{id}$ and $\alpha_{exp}$ may be inaccurate.

\subsubsection{Normalization using Other Face Models}
Apart from modeling 3D face structures with PCA, a number of approaches adopt other 3D modeling methods.
\citet{heo20093d} proposed an efficient approach called 3D Generic Elastic Models (GEM) for 3D modeling from a single frontal face image.
The fundamental assumption of GEM is that the depth information of human faces does not change dramatically between subjects.
It can be approximated from a generic depth map, as long as the frontal face and the depth map are densely aligned.
To establish the dense correspondence, 79 facial landmarks are first detected for both the frontal face and the depth map.
Delaunay Triangulation is utilized to obtain sparse 2D meshes from the landmarks.
To increase the vertex density of the meshes, loop subdivision is employed to generate new vertices using a weighted sum of the existing sparse vertices.
As the meshes for the frontal face and depth map are created in the same way, an accurate correspondence is built.
The depth value of the frontal mesh vertices is directly borrowed from those of the depth map,
while the texture value of the frontal mesh vertices remains the pixel value of the frontal image.
Thus, a textured 3D model of the frontal face is obtained.
\citet{prabhu2011unconstrained} applied GEM to PIFR.
They conducted 3D modeling for each frontal gallery image, and then generated non-frontal faces with the same pose as the probe face for matching.
GEM has the advantage in efficiency in the 3D modeling phase.
It also saves both the original shape and texture information of the frontal face, which is important for face recognition.
However, it applies only to the 3D modeling of frontal faces,
thus a number of non-frontal faces have to be rendered to extend the gallery set.

GEM is based on the strong approximation of the true depth of the face with a generic depth map.
To relieve this assumption,~\citet{heo2012gender} proposed the Gender and Ethnicity specific GEMs (GE-GEMs)
which employs gender and ethnicity specific depth maps for the 3D modeling of frontal faces.
The basic assumption of GE-GEMs is that the depth information of faces varies significantly less within the same gender and ethnicity group.
The authors empirically show that GE-GEMs can model the 3D shape of frontal faces more accurately than GEM.
The shortcoming of GE-GEMs is that it relies on correct gender and ethnicity classification in the testing time,
which is accomplished in a semi-automatic manner in this work.
Neither GEM or GE-GEM considers the influence of expression variation on the facial depth value.
\citet{Ali2015real} proposed the Probabilistic Facial Expression Recognition Generic Elastic Model (PFER-GEM)
to generate the image-specific depth map.
In PFER-GEM, the image-specific depth map is formulated as the linear combination of depth maps of four typical expressions,
therefore the constructed depth map is expression-adaptive.
The experimental results show that higher accuracy in 3D face modeling is achieved by PFER-GEM than GEM and GE-GEM.

\subsection{3D Pose Normalization from Multiple Images}
The approaches reviewed above attempt to conduct 3D modeling from a single input image,
since it is the most common setting for real-life face recognition applications.
The common shortcoming is that the personalized 3D shape parameters cannot be precisely approximated,
since 3D modeling from a single face image is essentially an ill-posed problem.
However, in some special applications, e.g., law enforcement, multi-view images are available for each subject during enrollment~\cite{zhang2006recognizing,zhang2008recognizing}.
In this case, it is desirable to utilize the multiple images together to build a more accurate 3D face model~\cite{xu2015multi}.

~\citet{zhang2005multilevel,zhang2008recognizing} proposed the Multilevel Quadratic Variation Minimization (MQVM)
approach to reconstruct an accurate 3D structure from a pair of frontal and profile face images of a subject.
~\citet{heo2011rapid} employed the depth information in the profile face image to modify the generic depth map in GEM.
In brief, a sparse 3D face shape is constructed by aligning the facial landmarks of the frontal and profile face images.
The depth map of the sparse face shape is merged with the generic depth map.
The combined depth map replaces the generic depth map in GEM for face synthesis.
It is argued that the combined depth map is more accurate and thus more realistic face images can be synthesized.
However, there are no experiments to validate the effectiveness of the proposed approach for face recognition.

Similarly, \citet{han20123d} extended the GEM approach to utilize the complementary information incorporated in the frontal and profile image pair.
Their approach is based on Eq.~\ref{E:pca3DShape} and Eq.~\ref{E:TAlphaOpt}, by which two 3D models are derived.
One is from the frontal face and the other is from the profile face, denoted as $S_{f}$ and $S_{p}$, respectively:
\begin{equation}
\begin{array}{cl}
S_{f}=\bar{S}+A \alpha_{f},\\
S_{p}=\bar{S}+A \alpha_{p}.
\end{array}
\label{E:hanNormalization}
\end{equation}
$S_{f}$ well preserves the 2D shape information of the frontal face,
while $S_{p}$ depicts more accurate depth information.
The information in $S_{f}$ and $S_{p}$ is complementary since they contain accurate 2D shape information and 3D depth information, respectively.
The final 3D shape model for the subject is obtained by replacing the depth value of 3D vertices in $S_{f}$ by that of $S_{p}$.
The 3D model is textured by mapping the pixel value in the frontal face image to the 3D model.

The pair of frontal and profile faces is not always available.
\citet{mostafa2012pose} proposed the reconstruction of a 3D structure of a subject during enrollment from stereo pair images,
captured by two cameras from arbitrarily different viewpoints.
The stereo pair images are employed to compute a disparity map using a stereo matching algorithm.
A cloud of 3D points is estimated from the disparity map and refined by surface fitting,
and a 3D triangular mesh is generated from the 3D points as the personalized 3D model for the subject.
For recognition, a set of images under discrete poses for each gallery subject is rendered using the 3D model and ray-tracing algorithm.
Given a probe image with non-frontal pose, the gallery images of the closest pose are employed for matching.
The experimental results highlight that using stereo-based 3D reconstruction
to synthesize images is more accurate than using a generic 3D shape model~\cite{mostafa2012pose,ali20143d}.

\subsection{3D Modeling by Image Reconstruction}
The 3D pose normalization approaches reviewed in the above two subsections take their cues from a set of facial landmarks for 3D shape reconstruction.
Compared to the whole face image, the information contained in the facial landmarks is limited,
which creates difficulty for accurate 3D face reconstruction and the subsequent face synthesis.
In contrast, the approaches reviewed in this subsection make full use of every pixel in the image to infer the 3D structure of the face image.

\citet{blanz1999morphable,blanz2003face} proposed the 3D Morphable Model (3DMM) approach to simulate the process of image
formation by combining a deformable 3D model and computer graphics techniques.
The deformable 3D model includes one shape vector and one texture vector.
The shape vector $S_{i}$ for the $i^{th}$ mesh is identical to Eq.~\ref{E:threeScan} and Eq.~\ref{E:pca3DShape}.
Similarly, the texture vector $T_{i}$ contains the color value of each vertex and is represented as follows,
\begin{equation}
\begin{array}{cl}
T_{i}=(r_{1}, g_{1}, b_{1}, r_{2},\dotsb, r_{n}, g_{n}, b_{n})^T\in R^{3n}.
\end{array}
\label{E:threeDTexture}
\end{equation}
The collection of the texture vectors for the $N$ scans is represented as $T = \left[T_{1}, T_{2}, \dotsb, T_{N}\right]\in R^{3n\times N}$.
By conducting PCA on $T$, the texture vector for a new 3D face $T_{t}$ is represented as
\begin{equation}
\begin{array}{cl}
T_{t}=\bar{T}+B \beta,
\end{array}
\label{E:pca3DTexture}
\end{equation}
where $\bar{T}$ is the mean texture of $T$. $B\in R^{3n\times k}$ is the matrix stacked of $k$ eigenvectors
and $\beta\in R^{k}$ contains the coefficients for the eigenvectors.

Computer graphics techniques employed in 3DMM incorporate the Phong illumination model~\cite{van2014computer} and the 3D-to-2D perspective projection model.
The illumination model accounts for the wide range of illumination variations in the face image, including cast shadows and specular reflections.
Given a single image, 3DMM automatically estimates the 3D shape coefficient vector $\alpha$, texture coefficient vector $\beta$,
and parameters of computer graphics models $\gamma$ by fitting the deformable 3D model with the face image.
The model parameters are optimized by a stochastic version of Newton's algorithm,
with the objective that the sum of squared differences over all pixels between the rendered image $I_{render}$ and the input image $I_{input}$ should be as similar as possible:
\begin{equation}
\begin{array}{cl}
&\min\limits_{\alpha,\beta,\gamma}\sum\limits_{x,y}\|I_{input}(x,y)-I_{render}(x,y)\|^{2}.
\end{array}
\label{E:3DMM}
\end{equation}

Ideally, the separated shape and texture parameters $\alpha$ and $\beta$ are only related to identity and thus provide pose and illumination invariance.
Face recognition can therefore be conducted by comparing $\alpha$ and $\beta$ between gallery and probe images.
Also, with the optimized $\alpha$ and $\beta$, 2D face images can be synthesized under arbitrary poses for the subject appearing in the input image.

The major disadvantage of 3DMM lies in its fitting procedure,
which is highly nonlinear and computationally expensive.
In practice, the contribution of shape, texture, pose, and illumination parameters to pixel intensity is ambiguous,
so the optimization of Eq.~\ref{E:3DMM} is non-convex and prone to be trapped into local minima.
Another disadvantage is that the PCA-based texture modelling may hardly represent the fine textures that are particular to individual faces~\cite{vetter1998synthesis}.

To reduce the computational complexity and relieve the non-convexity of 3DMM,
multiple constraints are employed to regularize the fitting procedure in Eq.~\ref{E:3DMM}.
For example,~\citet{romdhani2005estimating} proposed a new fitting algorithm that employs multiple features including pixel intensity, edges, and specular highlights.
They utilize a Bayesian framework and maximize the posterior of the shape and texture parameters given the multiple features.
The resulting cost function is smoother than Eq.~\ref{E:3DMM} and thus easier to optimize, rendering the fitting procedure more stable.
With the proposed fitting algorithm,
the authors reported higher face recognition accuracy and significantly better efficiency than the original fitting algorithm in 3DMM.
Other constraints incorporate facial symmetry prior~\cite{hu2013facial}, et al.

The non-convex optimization problem in Eq.~\ref{E:3DMM} can also be relieved with various multi-stage strategies.
For example,~\citep{kang2008multi,hu2012resolution} proposed the multi-resolution fitting methods that fit the 3DMM model
to the down-sampled low-resolution images and the original high-resolution image, successively.
~\citet{aldrian2013inverse} proposed the sequential estimation of shape and texture parameters.
First, the shape parameter $\alpha$ is estimated using Eq.~\ref{E:TAlphaOpt},
then the texture parameter $\beta$ and the illumination conditions are estimated by simulating the image formation process.
The optimization in each of the two steps is convex and thus can be solved with a global optimum.
Besides,~\citet{hu2014robust} proposed to first remove the illumination component of the input image using illumination normalization operations,
and then run 3DMM to optimize the shape, texture, and pose parameters only.

\subsection{Discussion}
3D pose normalization methods estimate only the pose and shape of a subject from one or multiple 2D face images,
while the texture information is directly mapped from the input 2D image to the 3D model.
The rendered images from the textured 3D model are consequently realistic in appearance.
Their shortcoming is that they usually make use of limited information, e.g., coordinates of dense facial landmarks,
to estimate the pose and shape parameters.
The error in pose and shape estimation results in undesirable artifacts in the subsequent texture mapping and face synthesis operations,
which adversely affects face recognition.
The 3D pose normalization methods also need special operations to handle the issue of missing data caused by self-occlusion.
In contrast, image reconstruction-based 3D modeling methods make full use of the textures appearing in the input 2D image.
The pose, shape, texture, and illumination parameters are estimated together by reconstructing the 2D image.
Face synthesis can be conducted with the estimated shape and texture parameters.
Their shortcoming lies in the difficulty in image fitting, which is often a non-convex problem,
thus the shape and texture parameters obtained may not be accurate,
which results in error in the synthesized face images.
As with the 2D face synthesis methods by regression models,
the synthesized images also lose detailed textures, e.g., moles, birthmarks, and wrinkles,
which is disadvantageous from the perspective of face recognition~\citep{park2010face,li2015high}.
Finally, one common disadvantage of 3D-based face synthesis methods is that they tend to lose information of the background surrounding the face region.
Recently,~\citet{zhu2015high} proposed to conduct 3D modeling of both the face and its background,
so that both information can be saved in the pose-normalized face image.

\section{Hybrid Methods}
The category of hybrid methods combines one or more of the aforementioned strategies for PIFR,
aiming to make use of the complementary advantages of these methods.
The hybrid approaches are less studied in the literature but tend to be more powerful than any single PIFR category,
and hold more promise for solving real-world PIFR problems.
Several successful combinations are reviewed below.

A number of approaches combine pose-robust feature extraction and multi-view subspace learning~\cite{prince2008tied,li2009maximizing,fischer2012analysis,li2012coupled}.
Instead of extracting holistic features from the whole face image,
they extract pose-robust features around facial landmarks,
substantially reducing the difficulty of multi-view subspace learning.
It has been shown that this strategy significantly enhances the performance of PIFR systems.
\citet{zhu2014recover} proposed a combination of 2D-based face synthesis and pose-robust feature extraction.
After obtaining the canonical view of a face image, the respective face component-level features are extracted for face matching.
\citet{ding2015multi} proposed the combination of 3D-based face synthesis and multi-view subspace learning,
inspired by the fact that frontal faces synthesized from non-frontal images of distinct poses differ in image quality,
which needs to be improved by multi-view subspace learning.

It is also possible to employ two or more categories of techniques independently and fuse their estimates into a single result.
For example, \citet{kim2006design} proposed an expert fusion system where pose-robust feature extraction expert, multi-view subspace learning expert,
and face synthesis expert run independently.
Results of the experts are fused in the score level and impressive performance improvement is observed.
Besides, it might be helpful to use 2D pose normalization and 3D pose normalization independently.
The 2D pose normalization methods are not good at correcting the nonlinear texture warping caused by pose variation,
but they can save all the information in the original 2D image,
e.g., facial contour, hairstyle, clothes, and backgrounds~\cite{berg2012tom,taigman2014deepface},
which are also important cues for face recognition~\cite{kumar2009attribute}.
Conversely, 3D pose normalization methods are good at correcting nonlinear texture warping
but tend to lose information outside of the facial area, as 3D face models only incorporate the facial area.
Combining the two face synthesis methods might be helpful for achieving a stronger PIFR system.
However, to the best of our knowledge, this fusion strategy has not been explored in the literature.

\section{Relations of the Four Categories}
In the above sections, we have discussed the pose-robust feature extraction methods, multi-view subspace learning methods,
and face synthesis-based methods independently.
Indeed, the three categories of methods try to solve the PIFR problem from different perspectives, as illustrated in Eq.~\ref{eq:generalFramework}.
In this section, we discuss the relative pros and cons of different categories.

High quality face representation is critical for the traditional NFFR problem,
and deep learning-based methods achieve great success in representation learning~\cite{taigman2015web,sun2015deeply,schroff2015facenet,ding2015robust}.
In the case of PIFR, we expect the pose-robust features extracted by powerful deep models continue to play a critical part,
provided the existence of massive labeled multi-pose training data.
However, multi-pose training data that is large enough to drive complicated deep models maybe difficult to collect in real-world applications.

It is not easy to independently exploit multi-view subspace learning methods because
they are based on the ideal assumption that simple pose-specific projections eliminate the cross-pose difference of faces.
Besides, their performance closely depends on the amount of labeled multi-pose training data.
In practice, multi-view subspace learning methods should be combined with pose-robust features,
which contribute to reducing the cross-pose gap.

Pose normalization-based face synthesis strategies are particularly successful when there is no multi-pose training data or the training data is small.
Their main limitation is the artifacts in the synthesized image caused by the inaccurate estimation of facial shape or pose parameters.
These artifacts change the original appearance of the subject,
and thus deteriorate the subsequently extracted features, causing an adverse impact on high-precision face recognition,
as empirically found in a recent work~\cite{ding2015robust}.

Therefore, both pose-robust feature extraction and pose normalization are promising solutions to PIFR.
In practice, the choice of the most appropriate PIFR method mainly depends on the availability and size of multi-pose training data.
Besides, the degree of pose variation is another important factor.
For near-frontal or half-profile face images, existing pose-robust feature extraction methods have already achieved high accuracy.
For profile face images, the synthesis-based approaches maybe useful to bridge the huge appearance gap between poses.
Moreover, hybrid methods try to solve the PIFR problem from multiple perspectives;
therefore they may be more promising to handle the real-world PIFR problem, as described in Section 7.

\section{Performance Evaluations}
\begin{table*}
\renewcommand{\arraystretch}{1.3}
\caption{Standard Datasets for Pose-Invariant Face Recognition}
\label{tab:Database}
\centering
\begin{tabular}{|p{13.5cm}|}
\hline

\textbf{FERET}  \ The multi-pose subset of the FERET database~\cite{phillips2000feret} contains 1,800 face images for 200 subjects across 9 poses.
                The 9 poses span $-65^\circ$ to $+65^\circ$ in yaw with $16^\circ$ intervals and there is no pitch variation.
                There is only one image for each subject under a certain pose.
                The database measures the influence of pose variation to face recognition exclusively, i.e.,
                other influential factors, e.g., illumination, expression, and background, remain the same across poses.
                \\
\textbf{CMU-PIE} \ The CMU Pose, Illumination, and Expression (PIE) database~\cite{sim2003cmu} contains 41,368 face images of 68 subjects across 13 poses, 43 illumination conditions, and 4 different expressions.
                The 13 pose types contain 9 poses with only yaw variation from the left profile to the right profile, with neighboring poses about $16^\circ$ apart.
                There are two poses containing pure pitch variations, and another two poses containing combined yaw and pitch variations.
                All images were captured in a single recording session.
                \\
\textbf{Multi-PIE} \ The Multi-PIE database~\cite{gross2010multi} contains images of 337 subjects from 15 different viewpoints,
                19 illumination conditions, and up to 6 expression types.
                Multi-PIE covers significantly more subjects than CMU-PIE,
                and the images of each subject were collected in up to four different recording sessions.
                There are 13 poses containing only yaw variations, ranging from $-90^\circ$ to $+90^\circ$, spaced in $15^\circ$ intervals.
                The remaining two poses contain hybrid yaw and pitch variations.
                \\
\textbf{CAS-PEAL-R1} \ The multi-pose subset of the CAS-PEAL-R1 database~\cite{gao2008cas} contains 21,832 face images of 1,040 subjects across 21 poses.
                All images were captured under ambient illumination and neutral expression. Each subject has only one image in each pose.
                The 21 poses are sampled from the pose space with 7 discrete yaw values and 3 discrete pitch values.
                The yaw values range from $-67^\circ$ to $+67^\circ$, while the pitch values are approximately $-30^\circ$, $0^\circ$, and $+30^\circ$.
                Images of the 21 poses are divided into three probe sets, namely PU, PM, and PD, each of which contains images of one particular pitch value.
                \\
\textbf{FacePix} \ The multi-pose subset of the FacePix database~\cite{little2005methodology}
                contains 181 images for each of the 30 subjects, covering yaw angle variations from $-90^\circ$ to $+90^\circ$, with an interval of $1^\circ$ between nearby poses.
                There are no pose variations in pitch and the facial expression stays neutral.
                The illumination condition is simulated ambient lighting and remains the same across poses.
                \\
\textbf{LFW}    \ The LFW database contains 13,233 images of 5,749 subjects, of which 1,680 subjects have two or more images.
                The images in LFW exhibit diverse variations in expression, illumination, and image quality that appear in daily life.
                However, faces in LFW are detected automatically by the simple Viola-Jones face detector~\cite{viola2004robust},
                which constrains the pose range of faces in LFW.
                In fact, the yaw values of more than 96\% of LFW images are within $\pm 30^\circ$,
                making LFW a less explored database for PIFR research.
                \\
\textbf{IJB-A}  \ The IARPA Janus Benchmark A (IJB-A) database~\cite{klare2015pushing} is a newly published face database
                which contains 5,712 face images and 2,085 videos from 500 subjects.
                Similar to LFW, images in IJB-A database are collected from Internet.
                The key characteristic of IJB-A is that both face detection and facial feature point detection are accomplished manually.
                Therefore, face images in IJB-A database cover full range of pose variations.
                \\
\hline
\end{tabular}
\end{table*}

\begin{table*}
\renewcommand{\arraystretch}{1.3}
\caption{Evaluation Summary of Different Categories of PIFR Algorithms on FERET}
\centering
\label{tab:FeretEva}
\begin{tabular}{llllll}
\hline
Publication  &\tabincell{l}{Pose\\Estimation}   &\tabincell{l}{Landmark\\Detection}
             &\tabincell{l}{Subject\\Number}    &\tabincell{l}{Pose Range\\(Yaw)}    &\tabincell{l}{Mean\\Accuracy}\\
\hline\hline
\cite{zhao2009textural}             &N/A        &N/A         &200      &$\pm 45^\circ$     &73.6 \\
\cite{arashloo2013efficient}        &N/A        &N/A         &200      &$\pm 65^\circ$     &97.4 \\
\cite{yi2013towards}                &Auto       &Auto        &200      &$\pm 65^\circ$     &97.4 \\
\cite{kan2014stacked}               &N/A        &Manual      &100      &$\pm 65^\circ$     &92.5 \\
\hline
\hline
\cite{li2009maximizing}             &Manual     &Manual      &100      &$\pm 65^\circ$     &81.5 \\
\cite{sharma2012robust}             &Manual     &Manual      &100      &$\pm 65^\circ$     &86.4 \\
\hline
\hline
\cite{ashraf2008learning}           &Manual     &Manual      &100      &$\pm 65^\circ$     &72.9 \\
\cite{li2012coupled}                &Manual     &Manual      &100      &$\pm 65^\circ$     &92.6 \\
\cite{li2012morphable}              &Manual     &Manual      &100      &$\pm 65^\circ$     &99.5 \\
\cite{li2014maximal}                &Manual     &Manual      &100      &$\pm 65^\circ$     &99.1 \\
\cite{ho2013pose}                   &N/A        &Auto        &200      &$\pm 45^\circ$     &95.5 \\
\hline
\hline
\cite{blanz2003face}                &Auto       &Manual      &194      &$\pm 65^\circ$     &95.8 \\
\cite{asthana2011fully}             &Auto       &Auto        &200      &$\pm 45^\circ$     &95.6 \\
\cite{mostafa2012dynamic}           &Auto       &Auto        &200      &$\pm 45^\circ$     &94.2 \\
\cite{Ali2015real}                  &Auto       &Auto        &200      &$\pm 65^\circ$     &99.1 \\
\hline
\hline
\cite{ding2015multi}                &Manual     &Manual      &100      &$\pm 65^\circ$     &99.6 \\
\hline
\end{tabular}
\begin{flushleft}
Mean accuracy is calculated based on the performance reported in the original papers.
Methods are organized in the order of pose-robust feature extraction, multi-view subspace learning,
2D-based face synthesis, 3D-based face synthesis, and hybrid categories.
\end{flushleft}
\end{table*}

\begin{table*}[!t]
\renewcommand{\arraystretch}{1.3}
\caption{Evaluation Summary of Different Categories of PIFR Algorithms on CMU-PIE}
\centering
\label{tab:PieEva}
\begin{tabular}{llllll}
\hline
Publication  &\tabincell{l}{Pose\\Estimation}         &\tabincell{l}{Landmark\\Detection}   &\tabincell{l}{Subject\\Number}
             &\tabincell{l}{Pose Range\\(Yaw/Pitch)}  &\tabincell{l}{Mean\\Accuracy}\\
\hline\hline
\cite{arashloo2011energy}       &N/A      &N/A         &68      &$\pm 65^\circ$ / $\pm 11^\circ$    &92.3 \\
\cite{yi2013towards}            &Auto     &Auto        &68      &$\pm 65^\circ$ / $\pm 11^\circ$    &95.3 \\
\hline\hline
\cite{prince2008tied}           &Manual   &Manual      &34      &$\pm 65^\circ$ / -                 &- \\
\cite{sharma2011bypassing}      &Manual   &Manual      &34      &$\pm 65^\circ$ / $\pm 11^\circ$    &93.9 \\
\hline\hline
\cite{chai2007locally}          &Manual   &Manual      &68      &$\pm 32^\circ$ / $\pm 11^\circ$    &94.6$^*$ \\
\cite{asthana2011pose}          &Auto     &Auto        &68      &$\pm 32^\circ$ / $\pm 11^\circ$    &95.0 \\
\cite{li2012coupled}            &Manual   &Manual      &34      &$\pm 65^\circ$ / $\pm 11^\circ$    &90.9 \\
\cite{li2012morphable}          &Manual   &Manual      &68      &$\pm 65^\circ$ / -                 &98.4$^*$ \\
\cite{li2014maximal}            &Manual   &Manual      &68      &$\pm 65^\circ$ / -                 &97.0$^*$ \\
\cite{ho2013pose}               &N/A      &Auto        &68      &$\pm 45^\circ$ / $\pm 11^\circ$    &98.8 \\
\hline\hline
\cite{asthana2011fully}         &Auto     &Auto        &67      &$\pm 32^\circ$ / $\pm 11^\circ$    &99.0 \\
\cite{mostafa2012dynamic}       &Auto     &Auto        &68      &$\pm 45^\circ$ / $\pm 11^\circ$    &99.3 \\
\cite{Ali2015real}              &Auto     &Auto        &68      &$\pm 65^\circ$ / $\pm 11^\circ$    &98.2 \\
\hline\hline
\cite{ding2015multi}            &Manual   &Manual      &68      &$\pm 65^\circ$ / $\pm 11^\circ$    &99.9$^*$ \\
\hline\hline
\end{tabular}
\begin{flushleft}
``$^*$'' indicates side information is utilized for model training.
Mean accuracy is calculated based on the performance reported in the original papers.
\end{flushleft}
\end{table*}

\begin{table*}[!t]
\renewcommand{\arraystretch}{1.3}
\caption{Summary of Representative PIFR Algorithms Evaluated on Multi-PIE}%
\centering
\label{tab:MultiPIEEva}
\begin{tabular}{lllll}
\hline
Publication             &Approach  &\tabincell{l}{Pose\\Estimation}   &\tabincell{l}{Landmark\\Detection}
                        &\tabincell{l}{Pose Range\\(Yaw/Pitch)}\\
\hline\hline
\cite{wright2009implicit}       &Implicit Matching     &N/A         &Auto        &$\pm 30^\circ$ / -\\
\cite{schroff2011pose}          &Doppelg{\"a}nger list &Manual      &Auto        &$\pm 90^\circ$ / $\pm 30^\circ$\\
\cite{zhangrandom}              &RF-SME                &N/A         &Manual      &$\pm 75^\circ$ / -\\
\cite{zhu2013deep}              &FIP+LDA               &N/A         &Manual      &$\pm 45^\circ$ / - \\
\cite{kafai2014reference}       &RFG                   &N/A         &N/A         &$\pm 90^\circ$ / - \\
\cite{kan2014stacked}           &SPAE                  &N/A         &Auto        &$\pm 45^\circ$ / -\\
\cite{zhu2014multi}             &MVP+LDA               &N/A         &Manual      &$\pm 60^\circ$ / - \\
\hline\hline
\cite{sharma2012robust}         &ADMCLS                &Manual      &Manual      &$\pm 90^\circ$ / $\pm 30^\circ$\\
\cite{sharma2012generalized}    &GMLDA                 &Manual      &Manual      &$\pm 75^\circ$ / -\\
\cite{kan2012multi}             &MvDA                  &Manual      &Manual      &$\pm 45^\circ$ / -\\
\hline\hline
\cite{li2012coupled}            &Ridge Regression      &Manual      &Manual      &$\pm 90^\circ$ / $\pm 30^\circ$\\
\cite{li2012morphable}         &SA-EGFC               &Auto        &Auto        &$\pm 45^\circ$ / - \\ %
\cite{li2014maximal}            &MDF-PM                &Manual      &Manual      &$\pm 90^\circ$ / - \\
\cite{ho2013pose}               &MRFs                  &N/A         &Auto        &$\pm 45^\circ$ / - \\
\cite{zhu2013deep}              &RL+LDA                &N/A         &Manual      &$\pm 45^\circ$ / - \\
\hline\hline
\cite{asthana2011fully}         &VAAM                  &Auto        &Auto        &$\pm 45^\circ$ / - \\
\cite{prabhu2011unconstrained}  &GEM                   &Auto        &Auto        &$\pm 60^\circ$ / - \\
\cite{heo2012gender}            &GE-GEM                &Manual      &Auto        &$\pm 30^\circ$ / - \\
\cite{zhu2015high}              &HPEN+LDA              &Auto        &Auto        &$\pm 45^\circ$ / - \\
\hline\hline
\cite{fischer2012analysis}      &Block Gabor+PLS       &Manual      &Manual      &$\pm 90^\circ$ / $\pm 30^\circ$\\
\cite{ding2015multi}            &PBPR-MtFTL            &Manual      &Manual      &$\pm 90^\circ$ / $\pm 30^\circ$\\
\hline\hline
\end{tabular}
\end{table*}

\begin{table*}[!t]
\renewcommand{\arraystretch}{1.3}
\caption{Identification Rates of PIFR Methods on Multi-PIE Using Protocol Defined in~\cite{asthana2011fully}}
\centering
\label{tab:ResultsMultiPIEICCV11}
\begin{tabular}{l|cccccc|c|c}
\hline
Publication                         &\tabincell{c}{080\\$-45^\circ$}  &\tabincell{c}{130\\$-30^\circ$}         &\tabincell{c}{140\\$-15^\circ$}  &\tabincell{c}{050\\$+15^\circ$}    &\tabincell{c}{041\\$+30^\circ$}    &\tabincell{c}{190\\$+45^\circ$}  &\tabincell{c}{Mean\\$\leq\pm45^\circ$}  &\tabincell{c}{Mean\\$\leq\pm90^\circ$}\\ \hline\hline
~\cite{zhu2013deep}                 &95.60  &98.50  &100.0  &99.30  &98.50  &97.80  &98.28  &-\\ 
~\cite{kan2014stacked}              &84.90  &92.60  &96.30  &95.70  &94.30  &84.40  &91.37  &-\\ 
~\cite{kafai2014reference}          &86.40  &91.20  &96.00  &96.10  &90.90  &85.40  &91.00  &-\\
~\cite{zhu2014multi}                &93.40  &100.0  &100.0  &100.0  &99.30  &95.60  &98.05  &-\\ \hline\hline
~\cite{kan2012multi}                &69.67  &83.33  &93.33  &93.00  &85.33  &69.33  &82.33  &-\\ \hline\hline
~\cite{li2012morphable}             &93.00  &98.70  &99.70  &99.70  &98.30  &93.60  &97.17  &-\\ 
~\cite{ho2013pose}                  &86.30  &89.70  &91.70  &91.00  &89.00  &85.70  &88.90  &-\\ 
~\cite{li2014maximal}               &90.00  &94.30  &95.30  &94.70  &93.70  &87.70  &92.62  &-\\ \hline\hline
~\cite{asthana2011fully}            &74.10  &91.00  &95.70  &95.70  &89.50  &74.80  &86.80  &-\\
~\cite{zhu2015high}                 &97.40  &99.50  &99.50  &99.70  &99.00  &96.70  &98.60  &-\\ \hline\hline
~\cite{ding2015multi}               &98.67  &100.0  &100.0  &100.0  &100.0  &98.33  &99.50  &92.04\\ \hline\hline
\end{tabular}
\begin{flushleft}
Performance of~\cite{kan2012multi} is obtained using the code released by the authors,
with DCP feature extracted from uniformly divided image regions as the face representation.
Performance can be improved when combined with pose-robust feature.
\end{flushleft}
\end{table*}

\begin{table*}[!t]
\renewcommand{\arraystretch}{1.3}
\caption{Summary of Representative PIFR Algorithms Evaluated on LFW}%
\centering
\label{tab:LFWEva}
\begin{tabular}{llllc}
\hline
Publication                      &Approach          &Category  &\tabincell{l}{Protocol}       &\tabincell{l}{Accuracy$\pm$ Std(\%)}\\
\hline\hline
\cite{arashloo2013efficient}     &MRF-MLBP          &Feature   &Protocol1        &79.08$\pm$0.14\\
\cite{li2015hierarchical}        &POP-PEP           &Feature   &Protocol1        &91.10$\pm$1.47\\
\cite{rahimzadeh2014class}       &MRF-Fusion-CSKDA  &Feature   &Protocol1        &95.89$\pm$1.94\\ \hline\hline
\cite{yi2013towards}             &PAF               &Feature   &Protocol2        &87.77$\pm$0.51\\
\cite{hassner2015effective}      &Sub-SML+LFW3D     &3D Synthesis   &Protocol2        &91.65$\pm$1.04\\
\cite{zhu2015high}               &HPEN+HD-Gabor+JB  &3D Synthesis   &Protocol2        &92.80$\pm$0.47\\ \hline\hline
\cite{chen2013blessing}          &High-dim LBP      &Feature   &Protocol3        &93.18$\pm$1.07\\
\cite{ding2014multi}             &MDML-DCPs         &Feature   &Protocol3        &95.58$\pm$0.34\\
\cite{ding2015multi}             &PBPR+PLDA         &3D Synthesis   &Protocol3        &92.95$\pm$0.37\\
\cite{zhu2015high}               &HPEN+HD-Gabor+JB  &3D Synthesis   &Protocol3        &95.25$\pm$0.36\\ \hline\hline
\cite{cao2010face}               &Multiple LE+comp  &Feature   &Protocol4        &84.45$\pm$0.46\\
\cite{chen2013blessing}          &High-dim LBP      &Feature   &Protocol4        &95.17$\pm$1.13\\
\cite{yin2011associate}          &Associate-Predict &2D Synthesis   &Protocol4   &90.57$\pm$0.56\\
\cite{berg2012tom}               &Tom-vs-Pete+Attribute &2D Synthesis   &Protocol4   &93.30$\pm$1.28\\
\cite{zhu2014recover}            &FR+FCN            &Hybrid   &Protocol4   &96.45$\pm$0.25\\
\hline\hline
\end{tabular}
\begin{flushleft}
``Feature'' stands for the pose-robust feature extraction category.
``2D Synthesis'' stands for the 2D-based face synthesis category.
``3D Synthesis'' stands for the 3D-based face synthesis category.
``Protocol1'' stands for the ``Image-Restricted, No Outside Data'' protocol~\cite{huang2014labeled}.
``Protocol2'' stands for the ``Image-Restricted, Label-Free Outside Data'' protocol.
``Protocol3'' stands for the ``Unrestricted, Label-Free Outside Data'' protocol.
``Protocol4'' stands for the ``Unrestricted, Labeled Outside Data'' protocol.
Results are directly cited from the original papers.
\end{flushleft}
\end{table*}

To compare the performance of different PIFR algorithms on a relatively fair basis,
a variety of face datasets have been established that range in scale and popularity.
Table~\ref{tab:Database} summarizes the existing datasets for PIFR research.
Of the existing databases, FERET~\cite{phillips2000feret}, CMU-PIE~\cite{sim2003cmu}, Multi-PIE~\cite{gross2010multi}, and LFW~\cite{LFWTech} are the most widely explored.
In the following, we conduct comparison and analysis of existing PIFR approaches based on their performance on the four databases.

\subsection{Evaluation on FERET and CMU-PIE}
The performances of representative PIFR algorithms on FERET and CMU-PIE are summarized in Table~\ref{tab:FeretEva} and Table~\ref{tab:PieEva}, respectively.
Since the vast majority of existing works conduct experiments on face identification,
we report their rank-1 identification rates as a metric for performance evaluation.
Note that the mean accuracy in Table~\ref{tab:FeretEva} and Table~\ref{tab:PieEva} is calculated based on the performance reported in the original papers.

The multi-pose subset of FERET and the CMU-PIE database are described in Table~\ref{tab:Database}.
For FERET, the frontal pose images are used as the gallery set, while all the non-frontal images are utilized as probe sets.
For CMU-PIE, methods in Table~\ref{tab:PieEva} use face images captured under neural expression and normal illumination conditions.
The frontal faces are utilized as the gallery set while the rest of the images compose the probe sets.
Note that some methods in Table~\ref{tab:FeretEva} and Table~\ref{tab:PieEva} need images of half subjects for training,
so the number of gallery subjects is 100 or 34 for these methods.

\citet{asthana2011fully},~\citet{li2012morphable}, and~\citet{ding2015multi}
reported the lowest error rates on FERET and CMU-PIE, demonstrating the power of 2D/3D pose normalization approaches.
This may be because pose normalization approaches correct the distortion caused by pose variation while saving the valuable texture detail.
The performance of these approaches has almost reached saturation point on FERET and CMU-PIE~\cite{ding2015multi}.
However, these results maybe unrealistically optimistic as they assume an ideal situation in which the illumination and expression conditions remain the same across poses.
It has been shown that the combined pose and illumination variations have a major influence
on the performance of PIFR algorithms on the CMU-PIE database~\cite{zhang2012heterogeneous,aldrian2013inverse,kafai2014reference}.

\subsection{Evaluation on Multi-PIE}
More recent approaches report recognition performance on the Multi-PIE database,
which covers images from more subjects and a wider range of pose, expression, and recording session variations.
The recent representative approaches that have conducted experiments on Multi-PIE are tabulated in Table~\ref{tab:MultiPIEEva}.

Several evaluation protocols on Multi-PIE exist, within which the protocol defined by~\citet{asthana2011fully} maybe the most popular one in the literature.
Under this protocol, images of the first 200 subjects are employed for training, and images of the remaining 137 subjects are used for testing.
Faces in the images have neutral expressions and frontal illumination, and were acquired across four recording sessions.
The frontal images from the earliest recording sessions for the testing subjects are collected as the gallery set.
The non-frontal images of the testing subjects constitute fourteen probe sets.
Performance of approaches that adopt this protocol is reported in Table~\ref{tab:ResultsMultiPIEICCV11}.
Among the methods in Table~\ref{tab:ResultsMultiPIEICCV11},
\citet{asthana2011fully} reported a mean accuracy of 86.8\% for probe sets whose yaw angles are within $\pm45^\circ$
by exploring the 3D pose normalization technique without occlusion detection.
Subsequently,~\citet{zhu2013deep} improved the performance to 95.6\% by learning pose-robust features via DNN.
\citet{ding2015multi} then achieved accuracy of 99.5\% by comprehensively employing 3D pose normalization,
image-specific occlusion detection, and multi-task discriminative subspace learning.

Considering that the protocol proposed by~\citet{asthana2011fully} is relatively simple,
\citet{zhu2014multi} proposed extending~\citet{asthana2011fully}'s protocol by incorporating probe images under all 20 illumination conditions.
The gallery images remain the same as those in~\citet{asthana2011fully}.
Similar to the experiments on CMU-PIE, the performance of the algorithms drops significantly under combined pose and illumination variations.
For example, the average performance of~\cite{zhu2013deep} and~\cite{ding2015multi} drop to 78.5\% and 96.5\% for probe sets within $\pm45^\circ$, respectively.

The robustness of PIFR algorithms to the combined pose and expression variations is less evaluated in the literature.
~\citep{schroff2011pose,kan2012multi,chu20143d,zhu2015high} provide the evaluations on the Multi-PIE database,
while \citet{jiang2005efficient} provide an evaluation on combined pose and expression variations on the CMU-PIE database.
The experimental results in~\cite{jiang2005efficient,chu20143d,zhu2015high} show that by synthesizing neutral and frontal face images with 3D normalization techniques,
the performance of traditional face recognition algorithms is significantly improved.

\subsection{Evaluation on LFW}
The above three most popular datasets for PIFR, i.e., FERET, CMU-PIE, and Multi-PIE, are small in scale and collected under laboratory conditions,
which means that they lack the diversity that appears in real-life faces.
Some recent PIFR algorithms evaluate their performance on the uncontrolled LFW database, which is composed of face images captured under practical scenarios.
However, it is important to note that LFW is actually designed for the NFFR task rather than PIFR,
since the yaw values of more than 96\% of LFW images are within $\pm 30^\circ$.

Performance of PIFR methods on LFW is tabulated in Table~\ref{tab:LFWEva}.
The works by~\cite{ding2014multi,rahimzadeh2014class,li2015hierarchical} highlight the importance of extracting pose-robust features.
In~\citet{ding2014multi}'s work, both holistic level and facial component level features are extracted from dense facial landmarks.
After fusing face representations from both levels, they achieve a $95.58\pm0.34\%$ accuracy under the ``Unrestricted, Label-Free Outside Data'' protocol.
In~\citet{rahimzadeh2014class}'s work,
semantically corresponding patches are densely sampled via MRF and features are extracted by multiple face descriptors.
By fusing a number of face representations for each image, they achieve a $95.89\pm 1.94\%$ accuracy under the ``Image-Restricted, No Outside Data'' protocol.

In comparison,~\cite{zhu2014recover,ding2015multi,hassner2015effective,zhu2015high}
show that various face synthesis techniques are helpful to promote the recognition performance on LFW.
For example, \citet{zhu2014recover} employed neural networks for 2D based face synthesis,
and reported a $96.45\pm0.25\%$ verification rate using the ``Unrestricted, Labeled Outside Data'' protocol.
In comparison,~\cite{ding2015multi,hassner2015effective,zhu2015high} utilized 3D pose normalization techniques for frontal face synthesis.
\citet{ding2015multi} reported a $92.95\pm0.37\%$ verification rate under the ``Unrestricted, Label-Free Outside Data'' protocol,
using only facial texture that is visible to both faces to be compared.
\citet{hassner2015effective} achieved a $91.65\pm1.04\%$ verification rate with the ``Image-Restricted, Label-Free Outside Data'' protocol.
With higher-fidelity 3D pose normalization operations, \citet{zhu2015high} reported $95.25\pm0.36\%$ accuracy under the ``Unrestricted, Label-Free Outside Data'' protocol.

It is clear that pose-robust feature extraction and face synthesis are two mainstream strategies to handle real-world pose variations.
Besides, as LFW is designed for the NFFR task rather than PIFR, new datasets that include full range of pose-variation images is desired to further test the performance of PIFR approaches.

\subsection{Efficiency Comparison of PIFR Algorithms}
In this subsection, the efficiency of some recent PIFR algorithms is briefly discussed,
so that readers can have a more comprehensive sense of the pros and cons of different approaches.
However, it is important to note that the algorithms are diverse in implementation.
Therefore, this subsection presents a qualitative comparison rather than a quantitative comparison.

Within pose-robust feature extraction methods,
the PAF approach proposed by \citet{yi2013towards} costs 70.9 ms (including facial landmark detection)
to re-establish the dense semantic correspondence between images on an Intel P4 CPU @ 2.0GHz,
and spends 18.7 ms on feature extraction at the semantically corresponding points.
In comparison, the MRF matching model by \citet{arashloo2013efficient} is computationally expensive.
It takes 1.4 seconds to establish the semantic correspondence between two images of size 112$\times$128 pixels on a NAVIDIA Gefore GTX 460 GPU,
while their previous work~\cite{arashloo2011energy} costs more than 5 minutes on a CPU for the same task.
Besides, the DNN-based FIP approach~\cite{zhu2013deep} spends 0.28 seconds to process one 64$\times$64 pixel image on an Intel Core i5 CPU @ 2.6GHz.

The multi-view subspace learning methods are generally fast in the testing time.
Their efficiency mainly differs in the training phase.
For example, the matlab implementation of the MvDA approach~\cite{kan2012multi}
takes about 120 seconds for model training on an Intel Core i5 CPU @ 2.6GHz,
with training data whose dimension is 600 from 4,347 images of 7 poses.

The efficiency of 2D-based face synthesis methods also differ significantly from each other.
Piece-wise-warping costs about 0.05 seconds using a single core Intel 2.2GHz CPU~\cite{taigman2014deepface}.
The MRF-based face synthesis method by~\citet{ho2013pose} takes less than two minutes
for frontal face synthesis from an input face image of size 130$\times$150 pixels, using an Intel Xeon CPU @ 2.13 GHz.

Within the 3D-based face synthesis methods, pose normalization based methods are generally efficient.
For example, pose normalization methods based on generic 3D face model cost about 0.1 seconds for frontal face synthesis from a 250$\times$250 pixel color image,
using matlab implementation on an Intel Core i5 CPU @ 2.6GHz~\cite{ding2015multi,hassner2015effective}.
The matlab implementation of the more complicated HPEN approach~\cite{zhu2015high} costs about 0.75 seconds for pose normalization on an Intel Core i5 CPU @ 2.6GHz.
In comparison, image reconstruction-based 3D face synthesis methods are computationally expensive.
For example, the classical 3DMM method takes 4.5 minutes on a workstation with an Intel P4 CPU @ 2GHz to fit the 3D face model and a 2D image~\cite{blanz2003face}.

\section{Summary and Concluding Remarks}
Pose-Invariant Face Recognition (PIFR) is the primary stumbling block to realizing the full potential of face recognition as a passive biometric technology.
This fundamental human ability poses a great challenge for computer vision systems.
The difficulty stems from the immense within-class appearance variations caused by pose change, e.g., self-occlusion, nonlinear texture distortion,
and coupled illumination or expression variations.

In this paper, we reviewed representative PIFR algorithms and classified them into four broad categories according to their strategy to bridge the cross-pose gap:
pose-robust feature extraction, multi-view subspace learning, face synthesis, and hybrid approaches.
The four categories of approach tackle the PIFR problem from distinct perspectives.
The pose-robust features can be grouped into two sub-categories: engineered features and learning-based features,
depending on whether the feature is extracted by manually designed descriptors or machine learning models.
The multi-view subspace learning approaches are divided into linear methods and nonlinear methods,
of which nonlinear methods show more promising performance.
The face synthesis category incorporates 2D-based face synthesis methods and 3D-based face synthesis methods,
depending on whether synthesis is conducted in the 3D domain or 2D domain.
Synthesis can be accomplished by simple pose normalization or various regression models,
and the normalization methods have the advantage of retaining the detail of facial textures.
Lastly, the hybrid methods combine two or more of these strategies for high performance PIFR.

Viewing the recent progress in PIFR, we have noticed some encouraging progress for each category of methods reviewed.
For example, extracting pose-robust features from semantic corresponding patches is becoming easier due to the rapid progress in facial landmark detection.
Researchers have also explored more advanced tools, e.g., 3D shape models, stereo matching, MRF, and GMM, to establish much denser semantic correspondence.
The recent explosion of nonlinear machine learning models, e.g., deep neural networks, are utilized to extract pose-robust features,
learn the multi-view subspace, and synthesize faces under novel poses.
Face synthesis based on 3D methods continues to be a hot topic because these methods work directly and explicitly to reduce the cross-pose gap in the 3D domain,
without requiring a large amount of multi-pose training data.
The 3D-based face synthesis approaches are still far from perfect,
and more accurate and stable algorithms are expected to be developed.
Another noticeable phenomenon is the increased number of hybrid approaches developed in recent years, which feature high PIFR performance.

Accompanying the development of PIFR algorithms is the introduction to the field of new databases, e.g., Multi-PIE and IJB-A,
which allow for more accurate evaluation to be conducted in more challenging environments.
However, images in Multi-PIE are collected under laboratory conditions and thus are not representative of realistic scenarios,
which may cause the PIFR task to be unrealistically easy.
The popular LFW database is designed for NFFR and contains a very limited number of profile or half-profile faces.
Fortunately, the newly introduced IJB-A database may fill the gap of unconstrained face database for PIFR research,
and developing larger scale unconstrained databases for PIFR is an emergent task.
Equally important is the reasonably designed evaluation protocol for each database, so that various algorithms can be directly compared.
Once such databases and corresponding evaluation protocols are available and being adopted, the real performance of existing PIFR algorithms in practical scenarios will be revealed,
and new insights for PIFR will be inspired.

Although great progress in PIFR has been achieved, there is still much room for improvement,
and the performance of existing approaches needs to be further evaluated on real-world databases.
To meet the requirement of practical face recognition applications, we propose the following design criteria
as a guide for future development.

\begin{itemize}
\item  \textbf{Fully automatic:}
       The PIFR algorithms should work autonomously, i.e., require no manual facial landmark annotations or pose estimation, etc.
       The recent progress of profile-to-profile landmark detection may enable this goal realized in the near future~\cite{xiong2015global}.
\item  \textbf{Full range of pose variation:}
       The PIFR algorithms should cover the full range of pose variations that might appear in the face image,
       including the yaw, pitch, and combined yaw and pitch.
       In particular, recognition of profile faces is very difficult and largely under-investigated.
       For pose normalization-based methods, the difficulty lies in the larger error of shape and pose estimation for profile faces.
       For pose-robust feature extraction-based methods, the difficulty is largely due to the lack of training data composed of labeled large-pose faces.
       Designing more advanced PIFR algorithms and collecting large-pose training data may be equally important.
\item  \textbf{Recognition from a single image:}
       The PIFR algorithms should be able to recognize a single non-frontal face utilizing a single gallery image per person.
       This is the most challenging but also the most common setting for real-world applications.
\item  \textbf{Robust to combined facial variations:}
       As explained in Section 1, the pose variation is often combined with illumination, expression, and image quality variations.
       A practical PIFR algorithm should also be robust to combined facial appearance variations.
       Existing pose normalization-based methods are sensitive to facial expression variations,
       due to the limitation of representation power of existing 3D face models.
       Therefore, more expressive 3D models that are competent to model non-rigid expression variations are required.
       Besides, the rich experience to handle uncontrolled illumination and image quality variations in the NFFR task
       may help tackle the more challenging PIFR task.
\item  \textbf{Matching between two arbitrary poses:}
       The most common setting for existing PIFR algorithms is to identify non-frontal probe faces from frontal gallery images.
       However, it is desirable to be able to match two face images with arbitrarily different poses, for both identification and verification tasks.
       One extreme example is to match a left-profile face to a right-profile face.
       In this case, the two faces have little visible regions in common.
       Using facial symmetry is an intuitive solution, i.e., turning the faces to the same direction~\cite{maurer1996single}.
       However, facial symmetry does not exactly hold true for high-resolution images where fine facial textures are clear.
       For low-resolution face images, e.g., frames of surveillance videos~\cite{Ross2015report}, this strategy may work well.
\item  \textbf{Reasonable requirement of labeled training data:}
       Although a large amount of labeled multi-pose training data helps to promote the performance of pose-robust feature extraction based PIFR algorithms,
       it is not necessarily available because labeled multi-pose data are difficult to collect in many practical applications.
       Possible solutions may incorporate making use of 3D shape priors of the human head and combining unsupervised learning algorithms~\cite{lecun2015deep}.
       Otherwise, methods that do not rely heavily on large amounts of training data, e.g., pose-normalization based methods, are advantageous.
\item  \textbf{Efficient:} The PIFR algorithms should be efficient enough to realize the requirement of practical applications,
       e.g., video surveillance and digital entertainment.
       Therefore, approaches that are free from complicated optimization operations in the testing time are preferable.
\end{itemize}

In the future, we would expect to see the evaluation of fully automatic and efficient PIFR algorithms on newly-developed real-life and large-scale face databases.
Since the existing PIFR approaches handle pose variations from distinct perspectives,
we would like to see individual improvement in pose-robust feature extraction, multi-view subspace learning, and face synthesis.
For example, it is anticipated that the engineered pose-robust features will be extracted from semantic corresponding patches both densely and efficiently.
The powerful nonlinear machine learning models, e.g., deep neural networks, are expected to be fully explored,
but at the cost of reasonable amount of multi-pose training data.
The face synthesis methods are expected to be able to accurately recover facial shape and textures under varied poses,
without artifact or statistical stability problems,
possibly by investigating more cues in the image or by combining several synthesis strategies.
We would also expect that the combination of several advanced techniques from multiple aspects, i.e., novel hybrid solutions,
will better accommodate the complex variations that appear in real images.

\bibliographystyle{spbasic}      

\end{document}